\documentclass[conference]{IEEEtran}
\IEEEoverridecommandlockouts
\usepackage{cite}
\usepackage{amsmath,amssymb,amsfonts}
\usepackage{algorithmic}
\usepackage{graphicx}
\usepackage{textcomp}
\usepackage{soul}
\usepackage{color,xcolor}

\usepackage[switch]{lineno}
\usepackage[ruled,linesnumbered]{algorithm2e}
\usepackage{subfigure}
\usepackage{multirow}
\usepackage{booktabs}
\usepackage{hyperref}
\usepackage{makecell}

\newcommand{\blue}[1]{\textcolor{blue}{#1}}

\def\BibTeX{{\rm B\kern-.05em{\sc i\kern-.025em b}\kern-.08em
		T\kern-.1667em\lower.7ex\hbox{E}\kern-.125emX}}
\begin{document}
	
	\title{Local-Global History-aware Contrastive Learning for Temporal Knowledge Graph Reasoning \\
	}
	
	\author{\IEEEauthorblockN{
			Wei Chen\IEEEauthorrefmark{2}\IEEEauthorrefmark{3}, 
			Huaiyu Wan\IEEEauthorrefmark{2}\IEEEauthorrefmark{3},
			Yuting Wu \IEEEauthorrefmark{4}\IEEEauthorrefmark{1},
			Shuyuan Zhao\IEEEauthorrefmark{2}\IEEEauthorrefmark{3},
			Jiayaqi Cheng\IEEEauthorrefmark{2},
			Yuxin Li\IEEEauthorrefmark{2},
			Youfang Lin\IEEEauthorrefmark{2}\IEEEauthorrefmark{3},
		}
		\IEEEauthorblockA{\textit{ \IEEEauthorrefmark{2}School of Computer and Information Technology, Beijing Jiaotong University, Beijing, China} \\
			\textit{ \IEEEauthorrefmark{4}School of Software Engineering, Beijing Jiaotong University, Beijing, China} \\ 
			\textit{ \IEEEauthorrefmark{3}Beijing Key Laboratory of Traffic Data Analysis and Mining, Beijing, China} \\ 
			\{w\_chen, hywan, ytwu1, sy\_zhao, jyq\_cheng, yuxinli, yflin\}@bjtu.edu.cn}\\
		\thanks{\IEEEauthorrefmark{1}Corresponding author: Yuting~Wu.}
	}
	
	\maketitle
	
	\begin{abstract}
		Temporal knowledge graphs (TKGs) have been identified as a promising approach to represent the dynamics of facts along the timeline. The extrapolation of TKG is to predict unknowable facts happening in the future, 
		holding significant practical value across diverse fields.
		Most extrapolation studies in TKGs focus on modeling global historical fact repeating and cyclic patterns, as well as local historical adjacent fact evolution patterns, showing promising performance in predicting future unknown facts.
		Yet, existing methods still face two major challenges: 
		(1) They usually neglect the importance of historical information in KG snapshots related to the queries when encoding the local and global historical information; 
		(2) They exhibit weak anti-noise capabilities, which hinders their performance when the inputs are contaminated with noise.
		To this end, we propose a novel \blue{Lo}cal-\blue{g}lobal history-aware \blue{C}ontrastive \blue{L}earning model (\blue{LogCL}) for TKG reasoning, which adopts contrastive learning to better guide the fusion of local and global historical information 
		and enhance the ability to resist interference.  
		Specifically, for the first challenge, LogCL proposes an entity-aware attention mechanism applied to the local and global  historical facts encoder, which captures the key historical information related to queries. For the latter issue,  LogCL designs a local-global query contrast module,  effectively improving the robustness of the model. 
		The experimental results on four benchmark datasets demonstrate that LogCL delivers better and more robust performance than the state-of-the-art baselines. 
		
	\end{abstract}
	
	
	\begin{figure}[htbp]
		\centering
		\includegraphics[width=1\linewidth]{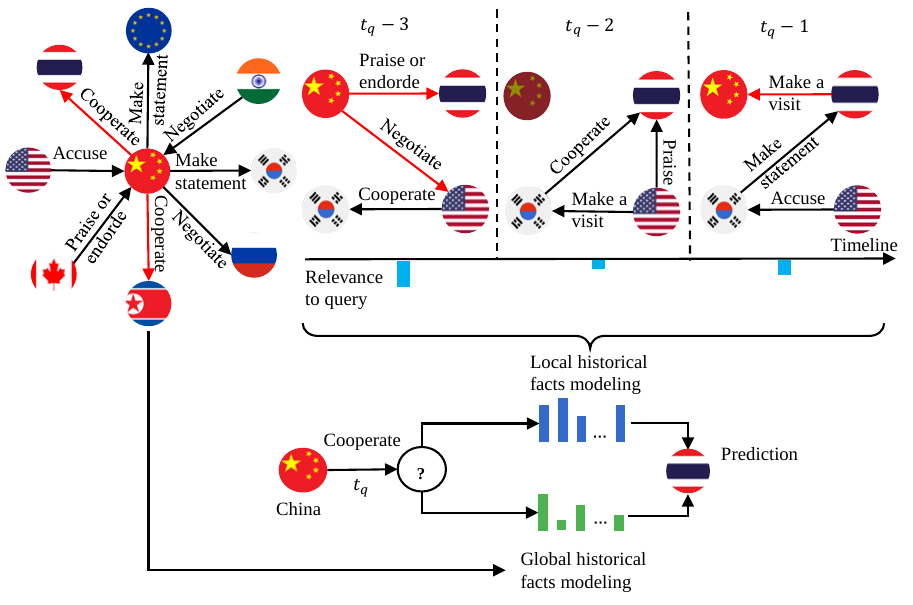}
		\caption{An illustrative example highlighting the importance of capturing the historical information related to the query. 
			The red arrows represent the most important facts associated with the query.
		}
		\label{fig1}
	\end{figure}
	\section{Introduction}
	Temporal knowledge graphs (TKGs), which represent dynamic facts as quadruples in the form of (subject, relation, object, time),  are actually sequences of KG snapshots with respective timestamps.  Reasoning on TKGs aims to predict the unknown facts by modeling the historical KGs snapshots, which involves two reasoning settings: interpolation and extrapolation. The interpolation setting focuses on completing the missing facts in history while the extrapolation setting aims to predict the facts happening in the future.  For TKGs, the extrapolation task is much more challenging and has great practical significance for various downstream applications, such as medical aided diagnosis system \cite{KnowLife,CuiSTMWL20}, traffic flow prediction \cite{Traffic} and stock prediction \cite{Stock}. Therefore, this paper focuses on extrapolation tasks that explore future unknown facts forecasting.
	
	Accurate predictions of future facts require a comprehensive understanding of the patterns of development of historical facts. According to human cognition of the development of historical facts\cite{trompf1979idea,evans1984heuristic,sloman1996empirical}, 
	predicting future facts involves the exploration of two historical  patterns: the repetition or cycling of historical facts  and the evolution of recent adjacent facts.
	Many efforts for TKG extrapolation have been made toward learning the facts repeating and cyclic patterns by global historical information and modeling the adjacent fact evolution patterns by local historical information. 
	
	The studies of the historical fact repeating or cyclic pattern such as CyGNet \cite{CyGNet}, aims to extract global repetition historical information for different queries in a heuristic way. This method leads to the narrow results that the predictions often lean towards the most frequently occurring facts. 
	The research on historical adjacent fact evolution pattern focuses on local historical facts temporal dependency modeling such as RE-GCN\cite{RE-GCN} and TANGO-Tucker\cite{TANGO}, but lacks the capture of global historical information. 
	Recently, some methods have tried to consider capturing both global and local historical patterns, such as TiRGN\cite{TiRGN} and HIPNet\cite{HIPNET}, which integrate global and local final prediction results to restrict the prediction range and achieve promising performance.  Yet, the performance of these methods is limited due to the following two challenges:
	
	\begin{figure}[htbp]
		\centering
		\subfigure[ICEWS14]{
			\begin{minipage}[t]{0.47\linewidth}
				\centering
				\includegraphics[width=1.05\linewidth]{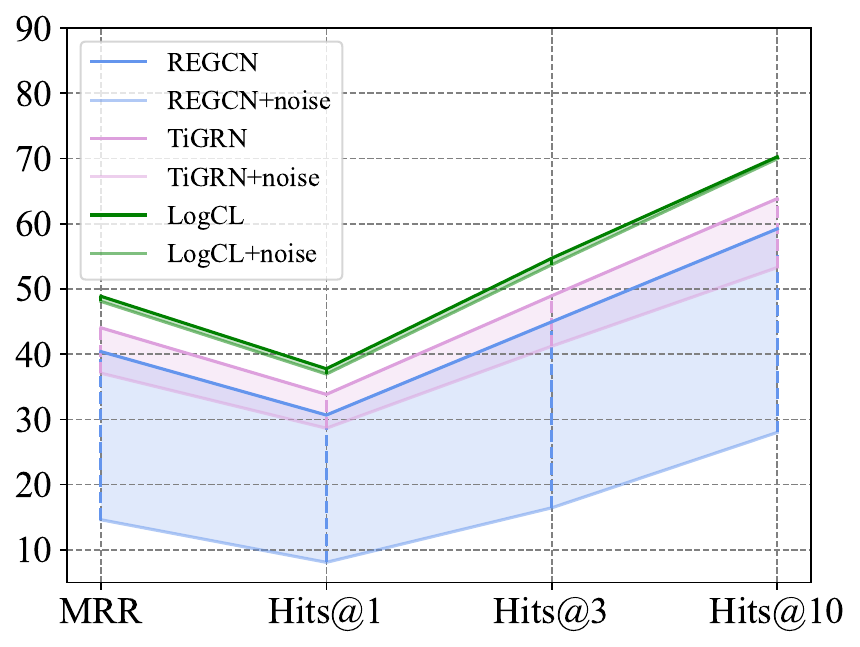}
			\end{minipage}
		}%
		\subfigure[ICEWS18]{
			\begin{minipage}[t]{0.47\linewidth}
				\centering
				\includegraphics[width=1.05\linewidth]{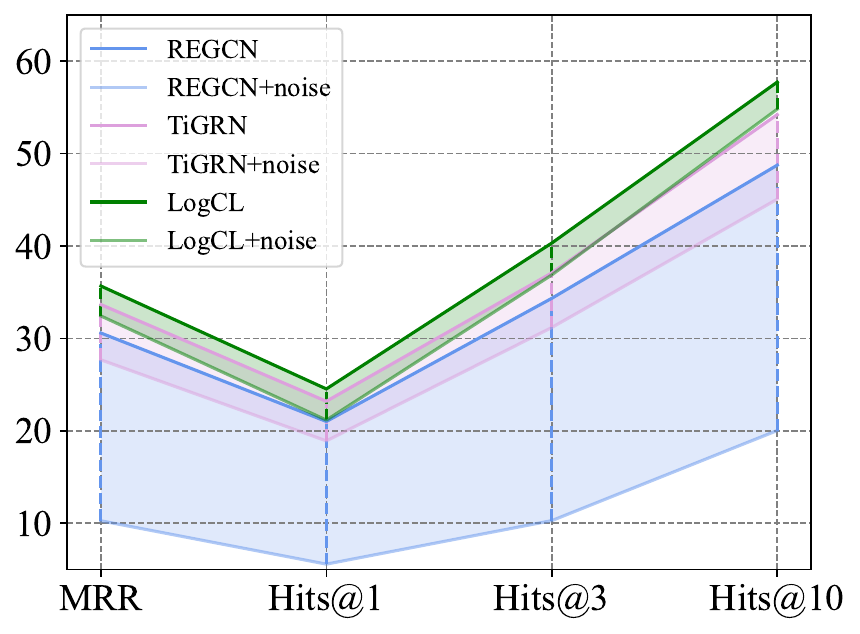}
			\end{minipage}
		}
		\caption{The Comparison results of REGCN\cite{RE-GCN}, TiRGN\cite{TiRGN} and our LogCL change after adding gaussian noise on ICEWS14 and ICEWS18 datasets. Light green shading, light red shading, and light blue shading indicate the variation range of LogCL, TiRGN, and REGCN, respectively. }
		\label{Fig2}
	\end{figure}

	\textbf{The importance of historical information related to the query is neglected during the process of encoding historical facts.}
	Most extrapolation methods on TKGs model the evolution of entities and relations by the order of time and argue that the KG snapshot facts closer to the query time are more important to predict the query. However, the entities in the query may not appear in all historical KG snapshots, which results in each snapshot taking on a different role in predicting the query. As shown in Fig. 1, for the query $(China, Cooperate, ?, t_q)$, the entity $China$ does not appear and is not directly related to other facts in $t_{q}-2$ KG snapshot, which provides little help for the prediction of the query. The facts containing the $China$ entity  in $t_{q}-2$ KG snapshot  appear earlier but are more helpful for predicting the query than $t_{q}-2$ KG snapshot.
	With the example in Fig. 1, we find that the historical facts in each KG snapshot are not always crucial for prediction.
	Existing approaches lack  essential patterns for capturing the important historical facts related to queries in the KG snapshot, impairing the ability to accurately predict future facts. So finding a way to filter the irrelevant KG snapshots based on the queries is much critical to improve the performance of TKG reasoning.
	
	\textbf{The weak anti-noise ability exists in existing methods that cannot effectively guarantee accurate predictions when the inputs adjoint noise.}  
	Mainstream TKG reasoning methods focus on how to model historical information and improve the accuracy of predictions. Nevertheless, the robustness of the model as a key important property to ensure the correct prediction results is rarely considered for TKG reasoning. In practice, during the process of training, the input data may encounter different noise interference, which leads to a significant degradation in prediction performance and even generates wrong prediction results. To further elaborate on the challenge mentioned above, we conduct experiments to simulate scenarios where the inputs are perturbed by adding Gaussian noise on both the classical  RE-GCN model and TiRGN model. The results shown in Fig. 2, it is  observed that both REGCN and TiRGN models suffer varying degrees of performance degradation when Gaussian noise is added, especially the MRR of REGCN model on ICEWS14 and ICEWS18 datasets are reduced by 63.8\% and 66.4\%, respectively. This indicates that the existing TKG method is weak in resisting the interference of noise. Therefore, how to effectively improve the robustness of the model to ensure correct prediction results is of great significance.
	
	To address the above challenges, we propose a novel \textbf{Lo}cal-\textbf{g}lobal history-aware \textbf{C}ontrastive \textbf{L}earning (LogCL) based on encoder-decoder structure for TKG reasoning, which  leverages contrastive learning to guide the fusion of local and global historical information and effectively enhance the robustness of the model. Specifically, during the model encoding phase, we propose an entity-aware  attention to flexibly learn query-related local and global historical information, thus forming a local entity-aware attention recurrent encoder and a global entity-aware attention encoder (First challenge). Inspired by the unsupervised contrastive learning \cite{Sup}, we design a local-global query contrast module to alleviate the interference brought by external noise to the model, which greatly improves the anti-noise ability of the model (Second challenge). As shown in Fig. 2, LogCL  performs better than REGCN and TiRGN  in the face of noise interference. In the final decoding part, we achieve entity prediction by fusing local and global historical information.
	
	In general, this work presents the following contributions:
	\begin{itemize}
		\item We propose a novel TKG extrapolation model that uses contrastive learning to better guide the fusion of 
		global and local historical information, 
		and enhance the robustness of LogCL. To the best of our knowledge, LogCL is the first model to leverage local-global history-aware contrastive learning to improve model robustness in TKG reasoning.
		\item We propose entity-aware attention applied to encode
		local and global historical information in an elegant way, which captures the importance of historical facts related to queries in KG snapshots.
		\item Extensive experiments on four public  datasets demonstrate that our proposed method shows better and more robust performance than
		the state-of-the-art baselines.
	\end{itemize}
	The rest of this paper is organized as follows. We first review the related work in Section II. Then we  describe the notation and problem statement, and introduce our proposed LogCL model  in Section III. Next, we report and analyze the experimental results in Section IV.  Finally, Section V concludes our paper.
	
	\section{Related Work}
	In this section, a review of the related work is presented, which includes KG reasoning methods and contrastive learning.
	\subsection{Reasoning on KGs}
	Reasoning on KGs has been a hot research topic that infers unknown facts based on known facts.
	Existing reasoning methods over KGs can be divided into two categories: static KG (SKG) reasoning and temporal KG reasoning. 
	\subsubsection{Static KG reasoning}
	Reasoning on SKG focus on static knowledge modeling, which can be roughly divided into four categories \cite{KGE, DRGI, UniKG}: translation-based methods, matrix factorization methods,  convolutional neural network (CNN)-based methods and graph neural network (GNN)-based methods. 
	Translation-based methods such as TransE \cite{TransE} and TransH \cite{TransH}, which learn embedded representations of entities and relations in a low-dimensional vector space. Matrix factorization methods including DistMult \cite{DisMult} and CompIEx \cite{CompIEx}, which adopt tensor/matrix factorization to obtain the latent representation of entities and relations. CNN-based methods such as ConvE \cite{Conv2D} and Conv-TransE \cite{Conv_TransE}, which  utilize CNNs to capture the interactions between entities and relations. GNN-based methods, such as R-GCN \cite{R-GCN}, CompGCN \cite{CompGCN} and KBGAT \cite{KBGAT}, which utilize GNNs to capture the structured semantic features of graphs. SKG reasoning methods have high efficiency in processing static knowledge. However, SKG reasoning methods do not take temporal information into consideration, which makes it difficult to model dynamic facts.
	\subsubsection{Temporal KG reasoning}
	Reasoning over TKGs involves two settings: interpolation and extrapolation. 
	The interpolation of TKGs infers the missing facts at historical timestamps. Most research works are based on the extension of SKG reasoning methods. For example, TTransE \cite{TTransE} extends the scoring function based on the idea of TransE\cite{TransE} and adds temporal constraints between facts that share entities. 
	HyTE\cite{HyTE} considers timestamps as time-related hyperplanes and generates time-aware representations by   projecting the entities and relations into time-related hyperplanes. 
	TNTComplEx\cite{TComplEX} extends the idea of tensor factorization and proposes a 4th-order tensor factorization to obtain the time-aware representations of entities. 
	Since these methods do not involve the evolution of facts over time, they are not suitable for modeling the entity representations in future unseen timestamps.
	
	TKG extrapolation aims to predict facts at future timestamps. 
	According to the law of human development of history cognition mainly involves two historical patterns:  the repetition or cycling of historical facts and the evolution of recent adjacent facts. The former \cite{CyGNet,CENET} explores the repetition of facts related to queries from a global perspective. For example, CyGNet \cite{CyGNet} and CENET \cite{CENET} adopt a copy-generation  mechanism to obtain the global repetition of one-hop facts, facilitating future entity reasoning. 
	The latter \cite{RE-NET,TANGO,TITer, XERTE, RE-GCN,CEN} aims to study the local historical facts temporal dependency. For example, RE-GCN \cite{RE-GCN} models the evolution of entities and relations at each timestamp by using the local historical dependency. CEN \cite{CEN} utilizes online learning strategies to deal with time-variability issues. GHT \cite{GHT} designs two Transformer modules \cite{Self-Attention} to capture instantaneous structural information and temporal evolution information, respectively, and presents a novel relational continuous-time encoding function to facilitate feature evolution with Hawkes processes.
	TECHS \cite{TECHS} encodes the topology and temporal dynamics by using heterogeneous attention and leverages both propositional and first-order reasoning to make predictions.  Recent methods such as TiGRN \cite{TiRGN} and HIPNet \cite{HIPNET}, attempt to integrate local and global historical information to obtain more accurate results. 
	Nevertheless, these approaches fail to model the importance of historical facts related to queries in KG snapshots. This makes it difficult to distinguish the significance of each KG snapshot for predicting the query.  To address this dilemma, we propose an entity-aware attention mechanism to flexibly model the KG snapshots that are relevant and irrelevant to the queries. 
	
	\subsection{Contrastive Learning}
	Contrastive Learning as a self-supervised learning paradigm, due to  its impressive performance in distinguishing instances of different categories,  has gained significant attention  in recent years.  The basic idea of contrastive learning \cite{SimpleCL,Sup} is to learn the similarity between different views of the same instances through training, while maximizing the dissimilarity between different views of instances. In self-supervised contrastive learning, the augmented instances are derived by randomly sampling a mini-batch of $N$ instances. The origin instances and the augmented instances are used to optimize the following loss function given a pair of positive instances $(i,j)$. The contrastive loss is expressed as:
	\begin{equation}
		\mathcal{L}_{i, j}=-\log \frac{\exp \left(\mathbf{x}_i \cdot \mathbf{x}_j / \tau\right)}{\sum_{k=1, k \neq i}^{2 N} \exp \left(\mathbf{x}_i \cdot \mathbf{x}_k / \tau\right)},
	\end{equation}
	where $2N$ is the sum of the number of the original and augmented instances, $\mathbf{x}_i$ is the projection embedding of the instance $i$, $\tau$ is a temperature parameter helping the model learn from hard negatives, and $\cdot$ is dot product that is used to compute the similarity of instances between different views. Recently, contrastive learning has been applied to static KG reasoning to alleviate the long-tail data problem but is rarely explored in TKG reasoning. The recent CENET \cite{CENET} is a single-view historical contrastive learning method designed to enhance the representation of entities with less historical interactions. Different from CENET, we propose a local-global query contrast module to mitigate noise interference and enhance model robustness.
	
	\begin{figure*}[htbp]
		\centering
		\includegraphics[width=0.8\linewidth]{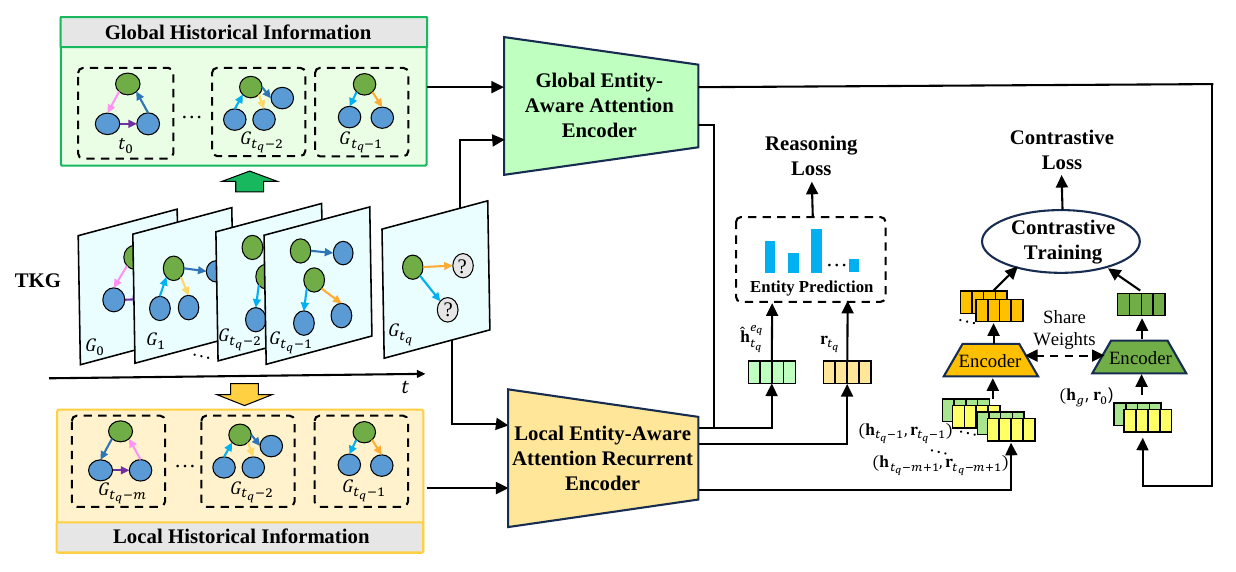}
		\caption{The overall  architecture of LogCL, consists of three components: local entity-aware attention recurrent encoder, global entity-aware attention historical encoder, and local-global query contrast module.  }
		\label{fig3}
	\end{figure*}
	
	\section{Our Approach}
	In this section, we will first introduce basic notations and definitions used in this paper and present the problem statement for the extrapolation of TKG, and then describe the framework and training optimizations of LogCL in detail.
	
	\begin{table}[h]
		\caption{\textbf{Summary of Primary Notations. }}
		\setlength\tabcolsep{3pt}
		\renewcommand{\arraystretch}{1.1}
		\centering
		\scalebox{1.05}{
			\begin{tabular}{ll}
				\toprule
				Notations &Descriptions       \\
				\midrule
				$\mathcal{G} $ &  A TKG   \\
				$\mathcal{G}_t$ & A KG snapshot composed of facts  at $t$ \\
				$\mathcal{G}_g$ & Global historical query subgraph \\
				$\mathcal{F}_t$ & Set of valid facts at  $t$\\
				$\mathcal{E}, \mathcal{R}, \mathcal{T}$ & Set of entities, relations and times in $\mathcal{G}$\\$|\mathcal{E}|,|\mathcal{R}|,|\mathcal{T}|$ & The number of entities, relations and times\\
				$d$ & Embedding dimensionality \\
				$m$ & Length of the local KG snapshot sequence \\
				$\mathbf{H}_0$ & Randomly initialized embedding matrices of entities \\
				$\mathbf{R}_0$ & Randomly initialized embedding matrices of realtions\\
				$ \mathbf{H}_{t}$ & Embedding matrices of entities at $t$   \\
				$\mathbf{R}_{t}$ & Embedding matrices of  relations at $t$\\
				$ \mathbf{H}_{g}$ & Global embedding matrices of entities    \\
				\bottomrule 
			\end{tabular}
			\label{table2}
		}
	\end{table}
	
	\subsection{Notations and Definitions}
	A TKG $\mathcal{G} $ is formalized as a sequence of KG snapshots, i.e.,  $\mathcal{G} = \{ \mathcal{G}_1, \mathcal{G}_2, ..., \mathcal{G}_{|\mathcal{T}|} \}$. Each KG snapshot $\mathcal{G}_t = (\mathcal{E},\mathcal{R},\mathcal{F}_t)$ actually is a directed multi-relational graph that is a set of valid facts at $t$. A fact (or an event) is denoted as a quadruple ($e_s, r, e_o, t$) where the subject entity $e_s \in \mathcal{E}$ and the object entity $e_o \in \mathcal{E}$ is connected by a relation $r \in \mathcal{R}$ at time $t\in \mathcal{T}$.  The primary notations and their meanings used in this paper are described in Table \uppercase\expandafter{\romannumeral1}.  
	
	TKG extrapolation task involves forecasting a missing object entity (or subject entity ) given a query $(e_{q}, r_{q}, ?,t_q)$ (or $(?, r_{q}, e_{q},t_q)$) according to previous historical KG snapshots $\{\mathcal{G}_{0}, \mathcal{G}_{1},..., \mathcal{G}_{t_{q}-1}\}$. Without loss of generality, the inverse relation quadruples $(e_o, r^{-1}, e_s,t)$ are added to the TKG dataset. So the TKG extrapolation task can be reduced to object entities predictions.
	
	\subsection{Architecture Overview} 
	As shown in Fig. 3, the overall framework of LogCL is composed of three components: global entity-aware attention encoder, local entity-aware recurrent encoder and local-global query contrast module.
	Local entity-aware attention recurrent encoder adopts an entity-aware attention mechanism to capture the necessary information related to queries for prediction. If facts in KG snapshots at the latest timestamps contain semantic information relevant to the queries, these facts will help to improve the probability of entity prediction.
	Global entity-aware attention encoder takes into account the global historical facts related to queries and avoids overlooking important historical facts that do not appear in the recent local KG snapshots.
	Local-global query contrast module is introduced to guide a better integration of global and local historical information and to enhance the robustness of LogCL. 
	Finally, LogCL combines local and global historical embedding representations for entity prediction.
	
	\begin{figure}[htbp]
		\centering
		\includegraphics[width=1.05\linewidth]{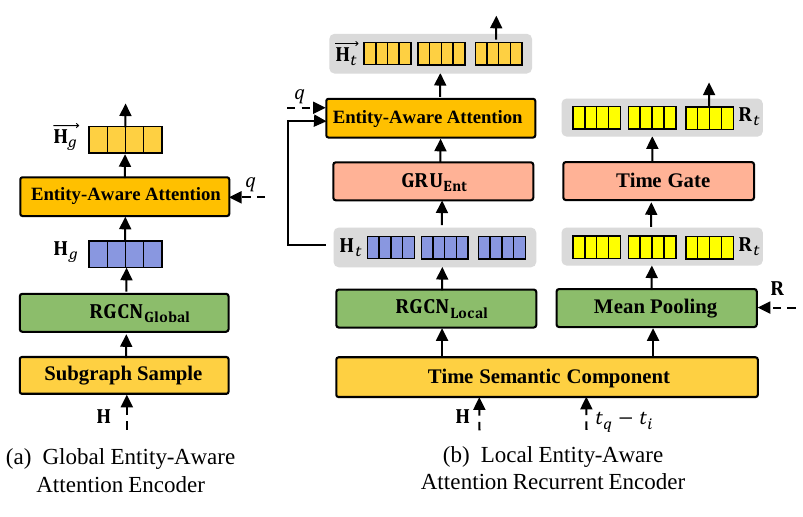}
		\caption{The architecture of our proposed  global entity-aware attention encoder and local entity-aware attention recurrent encoder.  }
		\label{fig4}
	\end{figure}
	
	\subsection{Local Entity-Aware Attention Recurrent Encoder}
	As discussed in the introduction, each KG snapshot  at the latest $m$ timestamp (i.e. $\{ \mathcal{G}_{t_q-m+1}, ..., \mathcal{G}_{t_q-1} \}$)  doesn't provide equally contributions for predicting the query $q=(e_{t_q}, r_{t_q}, ?, t_q)$. The importance of KG snapshot historical information relevant to the query needs to be modeled to further optimize the evolution of entity and relation in adjacent time. To this end, we propose a local entity-aware attention recurrent encoder that efficiently captures historical information in KG snapshots relevant to the query.  
	The local entity-aware attention recurrent encoder involves a KG snapshot aggregation pipeline and a KG snapshot sequences evolution pipeline. 
	Specifically,
	the aggregation of the KG snapshot is to learn the spatial semantic structure of concurrent facts by the Graph Convolution Network (GCN) from the spatial view, while the evolution of KG snapshot sequences is to capture sequential dependencies of entities and relations through the recurrent mechanism from the temporal view. Note that, the inputs to entities and relations on the first timestamp are randomly initialized.
	
	\subsubsection{KG Snapshot Aggregation}
	For KG snapshots on each timestamp, we  update the representation of entities by capturing  the spatial structural semantics information among concurrent facts. 
	Considering that some facts in KG snapshots occur cyclically, such as periodic meetings,  we first follow \cite{Hismatch} to encode the time numerical information to obtain the dynamic entities embedding. 
	Formally, the dynamic entities embedding as follows:
	\begin{equation}
		\resizebox{0.45\linewidth}{!}{
			\begin{math}
				\begin{aligned}
					\varphi(d)=\cos \left(d w_t+b_t\right),
				\end{aligned}
			\end{math}
		}
		\label{eq:2}
	\end{equation}
	\begin{equation}
		\resizebox{0.35\linewidth}{!}{
			\begin{math}
				\begin{aligned}
					\vec{\mathbf{h}}_t=W_0[\mathbf{h}_t \| \varphi(d)],
				\end{aligned}
			\end{math}
		}
		\label{eq:3}
	\end{equation}
	where $d = t_q-t_i$ is the time interval that is modeled by rescaling a learnable time unit $w_t$ with a time bias $b_t$. $\cos(\cdot)$ is a periodic activation function,  $\|$ is the vector concatenation operation, $W_0$ is a linear transformation matrix, and $\vec{\mathbf{h}}_t \in \vec{\mathbf{H}}_t$ is the dynamic embedding of entity at timestamp $t$.
	Then,  we utilize an R-GCN \cite{R-GCN} to capture the structure dependencies among concurrent facts. Note that, the R-GCN can be replaced by other
	relation-aware GCNs, such as CompGCN\cite{CompGCN} and KBGAT\cite{KBGAT}. The entity-aggregating R-GCN aggregator is defined as:
	\begin{equation}
		\resizebox{1\linewidth}{!}{
			\begin{math}
				\begin{aligned}
					& \vec{\mathbf{h}}_{t,o}^{(l+1)}= \mathrm{RGCN}_{Local}(\vec{\mathbf{h}}_{t,s}^{(l)},\mathbf{r}^{(l)},\vec{\mathbf{h}}_{t,o}^{(l)})\\
					&=\sigma_1 \left(\frac{1}{c_o} \sum_{(e_s, r), \exists(e_s, r, e_o) \in \mathcal{E}_t} \mathrm{~W}_1^{(l)}\left(\vec{\mathbf{h}}_{t,s}^{(l)}+\mathbf{r}^{(l)}\right)+\mathrm{W}_2^{(l)} \vec{\mathbf{h}}_{t,o}^{(l)}\right),
				\end{aligned}
			\end{math}
		}
		\label{eq:4}
	\end{equation}
	where $\vec{\mathbf{h}}_{t}^{l}\in \mathbb{R}^{|\mathcal{E}| \times d}, \mathbf{r}_{t}^{l} \in \mathbb{R}^{|\mathcal{R}| \times d}$ denote the embedding of entities and relations in the $l$-th layers of the R-GCN in each KG snapshot at $t$-th timestamp, separately. For simplicity, the final layer of the output of R-GCN is denoted as $\mathbf{H}_{t}^{Agg}$. ${c_o}$ is a normalization constant that equals the in-degree of entity,  $\mathrm{W}_{1}$ and $\mathrm{W}_{2}$ are the parameters for aggregating features and self-loop in the $l$-th layer, and $\sigma_1(\cdot)$ is the RReLU activation function. 
	\subsubsection{KG Snapshot Sequence Evolution}
	After obtaining the structural semantic embedding of entities in each KG snapshot, it needs to further model the sequential dependencies of entities and relations in KG snapshot sequences at the latest $m$ timestamps. We  progressively update the representations of entities by a gated recurrent unit (GRU) \cite{GRU} which is a flexible and efficient recurrent mechanism modeling short sequences. The  entity-oriented GRU can be represented as:
	\begin{equation}
		\resizebox{0.5\linewidth}{!}{
			\begin{math}
				\begin{aligned}
					\mathbf{H}_{t+1}=\operatorname{GRU_{\mathrm{Ent}}}\left(\mathbf{H}_{t}, \mathbf{H}_{t}^{Agg}\right), 
				\end{aligned}
			\end{math}
		}
		\label{eq:5}
	\end{equation}
	where $\mathbf{H}_{t}^{Agg}$ is the entity embedding matrix after KG snapshot aggeration at  $t$. As for relation, we follow \cite{RE-GCN} that considers $r$-related entities and the corresponding relation embedding to obtain $\mathbf{R}_{t}^{\prime}$  at  $t$ KG snapshot. Thus, a time gate unit is adopted to update the relation embedding at $t$. Formally, 
	\begin{equation}
		\resizebox{0.35\linewidth}{!}{
			\begin{math}
				\begin{aligned}
					\mathbf{r}_{t}^{\prime}=f_{ave}(\mathbf{H}_{t, r})+ \mathbf{r},
				\end{aligned}
			\end{math}
		}
		\label{eq:6}
	\end{equation}
	\begin{equation}
		\resizebox{0.56\linewidth}{!}{
			\begin{math}
				\begin{aligned}
					\mathbf{R}_{t+1}=\mathbf{U}_t \cdot \mathbf{R}_t^{\prime}+\left(1-\mathbf{U}_t\right) \cdot \mathbf{R}_{t},
				\end{aligned}
			\end{math}
		}
		\label{eq:7}
	\end{equation}
	\begin{equation}
		\resizebox{0.38\linewidth}{!}{
			\begin{math}
				\begin{aligned}
					\mathbf{U}_t=\sigma_2\left(\mathrm{W}_3 \mathbf{R}_t^{\prime}+\mathbf{b}\right),
				\end{aligned}
			\end{math}
		}
		\label{eq:8}
	\end{equation}
	where $f_{ave}(\cdot)$ is the mean pooling operation, $\mathbf{H}_{t, r}$ is the embedding of entities connected to $r$ at $t$, $\mathbf{R}_{t}^{\prime}$ consists of $\mathbf{r}_{t}^{\prime}$ at $t$, $\mathbf{R}_{t+1}$ is the finally updated through a time gate $\mathbf{U}_t \in \mathcal{R}^{d \times d}$, $W_3$ is the learnable weight matrice of the time gate, and $\sigma_2(\cdot)$ is the sigmoid activate function.
	
	After the above operations, we obtain the entity evolution representations $\{ \mathbf{H}_{t_q-m+2}, ..., \mathbf{H}_{t_q} \}$ and the relation evolution representations $\{ \mathbf{R}_{t_q-m+2}, ..., \mathbf{R}_{t_q} \}$ at the latest $m$ timestamps. However, these evolutionary representations do not take into account the contribution of historical facts related to queries in the KG snapshots for predicting the query. Thus, we propose an entity-aware attention mechanism to distinguish the importance of different KG snapshots for queries. We first exploit the mean pooling applied on all relation embedding associated with the entity in $q$ at timestamps $t_q$, and then compute the final local entity representation via the entity-aware attention mechanism. The operation is as follows,
	\begin{equation}
		\mathbf{h}_{t_q}^{e_q}=W_4[f_{ave}(\mathbf{r}_{t_q}) \| \mathbf{h}],
	\end{equation}
	\begin{equation}
		\resizebox{0.8\linewidth}{!}{
			\begin{math}
				\begin{aligned}
					\alpha_{i}= \sigma_2(W_5\left(\mathbf{h}_{t_q -m+i}^{Agg}+\mathbf{h}_{ t_q}^{e_q}\right)) \quad i \in[1, m-1], 
				\end{aligned}
			\end{math}
		}
		\label{eq:10}
	\end{equation}
	\begin{equation}
		\resizebox{0.5\linewidth}{!}{
			\begin{math}
				\begin{aligned}
					\vec{\mathbf{h}}_{t_q}^{e_q}= \mathbf{h}_{t_q} +\sum_{i=2}^{m} \alpha_i \mathbf{h}_{t_q-m+i},
				\end{aligned}
			\end{math}
		}
		\label{eq:11}
	\end{equation}
	where  $\alpha_i$ is the attention score of KG snapshots, $W_4$ and $W_5$ are the learnable weight matrices, $\sigma_2$ is the softmax acitvate function, and $\vec{\mathbf{h}}_{t_q}^{e_q}$ is the final local presentation at timestamp $t_q$ after local entity-aware attention recurrent encoder. In this way, LogCL can effectively model KG snapshot information related to queries at the latest $m$ timestamps, providing more accurate prediction results.
	
	\subsection{Global Entity-Aware Attention Encoder}
	The global entity-aware attention encoder aims to capture the longer historical facts that are not considered in the local KG snapshot sequence. For each query $(e_q,r_q,?,t_q)$,  unlike CyGNet \cite{CyGNet} and TiRGN\cite{TiRGN} that only use one-hop target object entities associated with queries to learn the repetition of facts in further history, we consider candidate multi-hop historical facts that provide more semantic information of historical facts to help entities prediction. For example, meetings that are held periodically are preceded by different hosting processes, and these different hosting processes can be of great help to future meetings.
	Therefore, we construct the historical query subgraph by sampling the historical facts relevant to queries to obtain rich semantic information.
	
	For a TKG  $\mathcal{G}_{<t_q}=\{ \mathcal{G}_1, \mathcal{G}_2, ..., \mathcal{G}_{t_q-1} \}$, we first sample the one-hop historical facts $\mathcal{G}^{\prime}_{g_1}$ containing the query subject entity $s$ in the given query. Next, we extract the one-hop target object entity associated with the query entity-relation pair, and then proceed to sample the one-hop facts $\mathcal{G}^{\prime}_{g_2}$ containing the one-hop target object entity. Finally, we integrate the two collected historical fact sets to obtain the most relevant historical subgraph to the query, $\mathcal{G}^{\prime}_{g} =\mathcal{G}^{\prime}_{g_1} \cup \mathcal{G}^{\prime}_{g_2} $. It is worth noting that, the historical query subgraph is a static KG that does not involve temporal information and can change along the query time.
	
	After obtaining the historical query subgraph, we update the global entity representation by modeling the structural semantic representation of the historical query subgraph. 
	To encode the historical query subgraph that is formed by sampling two-hop neighbors, we adopt another R-GCN to perform message aggregation. Since the historical query subgraph does not consider temporal information, we directly use the randomly initialized embedding of entities and relations as input to R-GCN. The specific formula is expressed as follows: 
	\begin{equation}
		\resizebox{0.6\linewidth}{!}{
			\begin{math}
				\begin{aligned}
					\mathbf{h}_{g}^{l+1} =\mathrm{RGCN}_{\mathrm{Global}}\left(\mathbf{h}_{g,e_s}^{l}, \mathbf{r}^{l},\mathbf{h}_{g,e_o}^{l}\right),
				\end{aligned}
			\end{math}
		}
		\label{eq:12}
	\end{equation}
	where $\mathbf{h}_{g}^{l+1} $ denotes the output embedding of entities in the $l$-th layers of the R-GCN in the historical query subgraph. For simplicity, the final layer of the output of R-GCN is denoted as $\mathbf{H}_{g}^{Agg}$. For global historical information, we also use the entity-aware attention mechanism to learn the historical facts related to queries. The specific formula is expressed as follows:
	\begin{equation}
		\resizebox{0.45\linewidth}{!}{
			\begin{math}
				\begin{aligned}
					\beta= \sigma_2(W_6\left(\mathbf{h}_{g}^{Agg}+\mathbf{h}\right)), 
				\end{aligned}
			\end{math}
		}
		\label{eq:13}
	\end{equation}
	\begin{equation}
		\resizebox{0.25\linewidth}{!}{
			\begin{math}
				\begin{aligned}
					\vec{\mathbf{h}}_{g}^{e_q}= \beta\mathbf{h}_{g}^{Agg},
				\end{aligned}
			\end{math}
		}
		\label{eq:14}
	\end{equation}
	where the $\beta$ is the attention score, $W_6$ is the learnable weight matrice, and $\vec{\mathbf{h}}_{g}^{e_q}$ denotes the final global representation of entities through global entity-aware attention encoder. 
	\subsection{Local and Global Historical Contrastive Learning}
	Clearly, the local and global encoders defined above well capture the local and global dependency for queries. 
	However, during the actual training process, the input data is disturbed by extraneous noise, leading to a sharp decline in reasoning performance.  Existing methods cannot address this challenge well. 
	Inspired by the unsupervised contrastive learning \cite{Sup,CENET}, we propose a local-global query contrast module that identifies highly correlated local and global features of entities and relations at the query level, enabling noise filtering. Specifically, for each query at timestamp $t_q$, 
	The purpose of the local-global query contrast module is to learn the local and global contrastive representations of queries by minimizing the supervised contrastive loss, which makes the local and global representations of the same query to be as close as possible in the semantic space, while different queries are separated. 
	Formally,  Formally,  the query embedding representations of local and global are obtained by using the concatenation operation:
	\begin{equation}
		\resizebox{0.35\linewidth}{!}{
			\begin{math}
				\begin{aligned}
					\mathbf{z}_{t}=MLP[\mathbf{h}_{t}^{Agg} \| \mathbf{r}_{t} ],
				\end{aligned}
			\end{math}
		}
		\label{eq:15}
	\end{equation}
	\begin{equation}
		\resizebox{0.35\linewidth}{!}{
			\begin{math}
				\begin{aligned}
					\mathbf{z}_{g}=MLP[\mathbf{h}_{g}^{Agg} \| \mathbf{r} ],
				\end{aligned}
			\end{math}
		}
		\label{eq:16}
	\end{equation}
	where $\mathbf{z}_{t}$ is the representation of the local query at timestamp $t$, $\mathbf{z}_{g}$ is the representation of the global query, and $MLP$ is used to normalize and project the embedding of queries onto the unit sphere for further contrastive training. It is worth noting that since the historical query subgraph does not consider temporal information, the origin relation embeddings are used to encode the global embedding representation of queries. 
	
	 In our work, the global and local query representations can be considered as the two-view representations derived from encoding historical facts in TKGs. Thus, if taking the query representation generated by the local encoder as the anchor, the global encoder can be viewed as an augmented encoding process. The local and global representations of the same query are used as positive pairs at timestamp $t$, i.e.,($\mathbf{z}_{t,i}$, $\mathbf{z}_{g,i}$), the local and global of representation of different queries are used as negative pairs i.e.,($\mathbf{z}_{t,i}$, $\mathbf{z}_{g,k}$). Formally, 
		the computation of the supervised contrastive loss $\mathcal{L}_{lg}$ at timestamp $t_q$ is as follows:
		\begin{equation}
			\mathcal{L}_{lg}= \frac{1}{|Q_{t_q}|}  \log \frac{\exp \left(\mathbf{z}_{t,i} \cdot \mathbf{z}_{g,i} / \tau\right)}{\sum_{k \in N_{t_q}, k \neq i }\left(\mathbf{z}_{t,i} \cdot \mathbf{z}_{g,k} / \tau\right)},
		\end{equation}
		where $Q_{t_q}$ and $N_{t_q}$ denote the minibatch that is the set of queries and the number of queries at timestamp $t_q$, separately; $\tau$ is the temperature parameter. The objective of $\mathcal{L}_{lg}$ is to make the representations of the same category closer, enhancing the common essential features of the local and global encoders required for predicting the query, thus alleviating the influence of noise and improving the robustness of the model. Similarly, the supervised loss $\mathcal{L}_{gl}$ can be obtained by using the representation generated by the global encoder as an anchor. 
		To make the LogCL  further distinguish different semantic representations of queries in the semantic space, we adopt Eq. 17 to constrain the local and global query representations and obtain two supervised losses $\mathcal{L}_{ll}$ and $\mathcal{L}_{gg}$.
		In this way, we can obtain the four supervised contrastive losses: $\mathcal{L}_{lg}$, $\mathcal{L}_{gl}$, $\mathcal{L}_{ll}$ and $\mathcal{L}_{gg}$.
		Thus, the final supervised contrastive loss is computed as $\mathcal{L}_{cl}= (\mathcal{L}_{lg}+\mathcal{L}_{gl}+\mathcal{L}_{ll}+\mathcal{L}_{gg})/4$.

	
	\subsection{Prediction and Optimization}
	ConvTransE\cite{Conv_TransE} is a strong score function that is widely adopted for the recent TKG reasoning task. In this work, we also adopt ConvTransE  to perform the entity prediction task at timestamp $t_q$. Therefore, the entity prediction score of ConvTransE is as follows:
	\begin{equation}
		\begin{aligned}
			\phi \left(e_{q}, r_{q}, e, q\right)= \sigma_2 \left(\mathbf{h}_{t_q}^{e_q} \operatorname{ConvTransE}\left(\hat{\mathbf{h}}_{t_q}^{e_q}, \mathbf{r}_{t_q}\right)\right),
		\end{aligned}
	\end{equation}
	\begin{equation}
		\hat{\mathbf{h}}_{t_q}^{e_q} = \lambda\vec{\mathbf{h}}_{g}^{e_q}+(1-\lambda)\vec{\mathbf{h}}_{t_q}^{e_q},
	\end{equation}
	where $\lambda$ is a variable factor that is set at [0,1], which is used to trade off global and local representations of entities. In this paper, the entity prediction task can be considered as a multi-label learning problem. Thus, the loss of entity prediction $\mathcal{L}_{tkg}$ is formalized as:
	\begin{equation}
		\begin{aligned}
			\mathcal{L}_{tkg} = \sum_{t=0}^\mathcal{T} \sum_{\left(e_s, r, e, t\right) \in \mathcal{F}_t} \sum_{e \in \mathcal{E}} y_t^e \log \phi\left(e_s, r, e, t\right),
		\end{aligned} 
	\end{equation}
	where $\phi\left(e_s, r, e, t\right)$ is the entity prediction probabilistic scores. $\mathbf{y}_{t}^e \in \mathbb{R}^{|\mathcal{E}|}$  is the label of which the element is 1 if the fact occurs, otherwise 0. The final loss function is computed as: 
	\begin{equation}
		\resizebox{0.32\linewidth}{!}{
			\begin{math}
				\begin{aligned}
					\mathcal{L} = \mathcal{L}_{tkg}+\mathcal{L}_{cl}.
				\end{aligned}
			\end{math}
		}
		\label{eq:21}
	\end{equation}
	
	In the work, the contrastive supervised loss $\mathcal{L}_{cl}$ and the entity prediction of loss $\mathcal{L}_{tkg} $ are trained simultaneously. 
	
	Note that, since reverse quadruples are added in the training or testing process, the inverse query sets are generated by the original query sets at timestamp $t_q$. During each epoch, the KG snapshots in history are built using the combination of the original quadruples and the inverse quadruples. If the combination sets are directly used for training, the entity-aware attention will perceive the target subject entity and target object entity, resulting in data leakage.
	To avoid such potential data leakage, we present a two-phase forward propagation strategy. In detail, the original query set is first considered in the forward propagation, and then the second forward propagation is performed on the query inverse sets during each epoch. In this way, the prediction scores and loss can be obtained for training or testing from the two-phase forward propagation. The detailed training procedure of LogCL is summarized in Algorithm 1.
	
	\begin{algorithm}[!h]
		\renewcommand{\algorithmicrequire}{\textbf{Input:}}
		\renewcommand{\algorithmicensure}{\textbf{Output:}}
		\caption{Training procedure of LogCL}
		\begin{algorithmic}[1]
			\REQUIRE the historical KG snapshot sequence \\$\{ \mathcal{G}_{1}, \mathcal{G}_{2}, ..., \mathcal{G}_{t_q} \}$, query set $\mathcal{Q}_{query}$ with unknown \\ object entities at time $t_q$.
			\ENSURE The reasoning results for each query are in descending order of scores.
			\STATE Initialize the embeddings of entities, relations.
			\WHILE {$t< |\mathcal{T}| $}
			\STATE $t' = t_q-m$
			\WHILE {$t^{\prime}<t_q$ and $t_q^{\prime}>0$}
			\STATE $t' = t'+1$
			\STATE Aggregate local KG snapshot by Eq.2-Eq.4.
			\STATE Learn KG snapshot sequence evolution and obtain the presentation of entities and relations by \\Eq.5-Eq.11.
			\ENDWHILE
			\STATE Aggregate global query subgraph and obtain the representation of entities by Eq.12-Eq.14.
			\STATE $t'' = t-m$
			\WHILE {$t^{\prime \prime}<t$ and $t^{\prime \prime}>0$}
			\STATE $t'' = t''+1$
			\STATE Compute the local-global query contrast loss by Eq.15-Eq.17.
			\ENDWHILE
			\STATE Replace the missing object entities for each query $(s, r, ?, t) \in \mathcal{Q}_{query}$  and calculate the scoring\\ function by Eq.18-Eq.21.
			\STATE $t = t+1$
			\ENDWHILE
		\end{algorithmic}
	\end{algorithm}
	
	\subsection{Complexity Analysis}
	We analyze the computational complexity of our proposed LogCL model. For the local entity-aware attention recurrent encoder, the time complexities of entity and relation evolution  are $O(m(|\mathcal{E}|+P))$ and $O(m|\mathcal{R}|)$,  respectively,  where $P$ is the complexity of entity attention mechanism at $t$ timestamp. For the global entity-aware encoder, the time complexity of generating the historical query subgraph is $O(2|\mathcal{T}||\mathcal{F}_{t}|)$, the time complexity of historical query subgraph aggregation is $O(|\mathcal{E}|+P)$. The time complexity of the local-global query contrast module is $O(4mC)$.  Thus, the time complexity of LogCL is  $O(m(|\mathcal{E}+\mathcal{R}+P|)+|\mathcal{E}|+P+2|\mathcal{T}||\mathcal{F}_{t}|+4mC)$.
	
	\begin{table}[h]
		\caption{\textbf{DETAILS OF THE TKG DATASETS. }}
		\setlength\tabcolsep{3pt}
		\renewcommand{\arraystretch}{1.1}
		\centering
		\scalebox{1.0}{
			\begin{tabular}{lllll}
				\toprule
				Dataset &ICEWS14  &ICEWS18 &ICEWS05-15 & GDELT     \\
				\midrule
				Entities &6,869  &10,094 &23,033 & 7,691  \\
				Relations & 230 & 256& 251 & 240\\
				Training & 74,845 & 373,018& 368,868& 1,734,399\\
				Validation & 8,514  &45,995 &46,302 &238,765\\
				Test & 7,371 &49,545 &46,159 &305,241\\
				Time granularity & 24 hours &24 hours &24 hours & 15 mins\\
				Snapshot numbers & 365 & 365 & 4017 & 2975   \\
				\bottomrule 
			\end{tabular}
			\label{table3}
		}
	\end{table}
	
	\begin{table*}[t]
		\caption{The prediction performance of MRR and Hits@1/3/10 are on all datasets with time-aware metrics.}
		\setlength\tabcolsep{2pt}
		\renewcommand{\arraystretch}{1.2}
		\centering
		\scalebox{0.93}{
			\begin{tabular}{llcccccccccccccccc}
				\toprule
				&\multirow{3}*{Model} 
				&\multicolumn{4}{c}{ICEWS14}&\multicolumn{4}{c}{ICEWS18}&\multicolumn{4}{c}{ICEWS05-15}&\multicolumn{4}{c}{GDELT} \\
				\cmidrule(lr){3-6} \cmidrule(lr){7-10} \cmidrule(lr){11-14} \cmidrule(lr){15-18}
				
				&&MRR &Hits@1 &Hits@3 &Hits@10 &MRR &Hits@1 &Hits@3 &Hits@10 &MRR &Hits@1 &Hits@3 &Hits@10 &MRR &Hits@1 &Hits@3 &Hits@10\\
				\midrule
				\multirow{4}{*}{\rotatebox{90}{Static}}
				&DisMult (2014) & 15.44  & 10.91  & 17.24  & 23.92  & 11.51  & 7.03  & 12.87  & 20.86  & 17.95  & 13.12  & 20.71  & 29.32  & 8.68  & 5.58  & 9.96  & 17.13\\
				&ComplEx (2016) & 32.54  & 23.43  & 36.13  & 50.73  & 22.94  & 15.19  & 27.05  & 42.11  & 32.63  & 24.01  & 37.50  & 52.81  & 16.96  & 11.25  & 19.52  & 32.35 \\
				&ConvE (2018) & 35.09  & 25.23  & 39.38  & 54.68  & 24.51  & 16.23  & 29.25  & 44.51  & 33.81  & 24.78  & 39.00  & 54.95  & 16.55  & 11.02  & 18.88  & 31.60  \\
				&Conv-TransE (2019)& 33.80  & 25.40  & 38.54  & 53.99  & 22.11  & 13.94  & 26.44  & 42.28  & 33.03  & 24.15  & 38.07  & 54.32  & 16.20  & 10.85  & 18.38  & 30.86  \\
				&RotatE (2019)& 21.31  & 10.26  & 24.35  & 44.75  & 12.78  & 4.01  & 14.89  & 31.91  & 24.71  & 13.22  & 29.04  & 48.16  & 13.45  & 6.95  & 14.09  & 25.99 \\
				\midrule
				\multirow{4}{*}{\rotatebox{90}{Interpolation}}
				&TTransE (2016) &13.72  & 2.98  & 17.70  & 35.74  & 8.31  & 1.92  & 8.56  & 21.89  & 15.57  & 4.80  & 19.24  & 38.29  & 5.50  & 0.47  & 4.94  & 15.25 \\
				&TA-DisMult (2018) & 25.80  & 16.94  & 29.74  & 42.99  & 16.75  & 8.61  & 18.41  & 33.59  & 24.31  & 14.58  & 27.92  & 44.21  & 12.00  & 5.76  & 12.94  & 23.54 \\ 
				&DE-SimIE (2020) & 33.36  & 24.85  & 37.15  & 49.82  & 19.30  & 11.53  & 21.86  & 34.80  & 35.02  & 25.91  & 38.99  & 52.75  & 19.70  & 12.22  & 21.39  & 33.70  \\
				&TNTCompIEx (2020) &34.05  & 25.08  & 38.50  & 50.92  & 21.23  & 13.28  & 24.02  & 36.91  & 27.54  & 9.52  & 30.80  & 42.86  & 19.53  & 12.41  & 20.75  & 33.42 \\
				\midrule
				\multirow{10}{*}{\rotatebox{90}{Extrapolation}}
				&RE-NET (2020) &36.93  & 26.83  & 39.51  & 54.78  & 28.81  & 19.05  & 32.44  & 47.51  & 43.32  & 33.43  & 47.77  & 63.06  & 19.62  & 12.42  & 21.00  & 34.01 \\
				&CyGNet (2020) & 35.05  & 25.73  & 39.01  & 53.55  & 24.93  & 15.90  & 28.28  & 42.61  & 36.81  & 26.61  & 41.63  & 56.22  & 18.48  & 11.52  & 19.57  & 31.98 \\
				&TANGO-Tucker (2021)& 36.80  & 27.43  & 40.89  & 54.93  & 28.68  & 19.35  & 32.17  & 47.04  & 42.86  & 32.72  & 48.14  & 62.34  & 19.53  & 12.43  & 20.79  & 33.19  \\
				&xERTE (2021) & 40.02  & 32.06  & 44.63  & 56.17  & 29.98  & 22.05  & 33.46  & 44.83  & 46.62  & 37.84  & 52.31  & 63.92  & 18.09  & 12.30  & 20.06  & 30.34 \\
				&TITer (2021) &  40.87  & 32.28  & 45.45  & 57.10  & 29.98  & 22.05  & 33.46  & 44.83 & 47.69  & 37.95  & 52.92  & 65.81  & 15.46  & 10.98  & 15.61  & 24.31 \\
				&RE-GCN (2021) & 40.39  & 30.66  & 44.96  & 59.21  & 30.58  & 21.01  & 34.34  & 48.75  & 48.03  & 37.33  & 53.85  & 68.27  & 19.64  & 12.42  & 20.90  & 33.69 \\
				&CEN (2022) & 42.20  & 32.08  & 47.46  & 61.31  & 31.50  & 21.70  & 35.44  & 50.59  & 46.84  & 36.38  & 52.45  & 67.01  & 20.39  & 12.96  & 21.77  & 34.97  \\
				&TiRGN (2022) & 44.04  & 33.83  & 48.95  & 63.84    & 33.66      &23.19       &\underline{37.99}       &\underline{54.22}       & 50.04     &39.25     &56.13     & 70.71 & 21.67      &13.63       &23.27       &\underline{37.60}\\
				&HisMatch (2022) & \underline{46.42}  & \underline{35.91}  & \underline{51.63}  & \underline{66.84}    &\underline{33.99}      &\underline{23.91}       &37.90       &53.94       & \underline{52.85}     &\underline{42.01}     &\underline{59.05}     & \underline{73.28} &\underline{22.01}      &\underline{14.45}   &\underline{23.80}       &36.61 \\
				&RETIA (2023) &42.76  & 32.28  & 47.77  & 62.75  & 32.43  & 22.23  & 36.48  & 52.94  & 47.26  & 36.64  & 52.90  & 67.76   & 20.12      &12.76      &21.45      &34.49   \\
				&CENET (2023)& 39.02 & 29.62 & 43.23 & 57.49 & 27.85 & 18.15 & 31.63 & 46.98 & 41.95 & 32.17 & 46.93 & 60.43 & 20.23 & 12.69 & 21.70 & 34.92 \\
				\midrule
				&LogCL &\textbf{48.87}  & \textbf{37.76}  & \textbf{54.71}  & \textbf{70.26}  & \textbf{35.67}  & \textbf{24.53}  & \textbf{40.32}  & \textbf{57.74}  & \textbf{ 57.04}  & \textbf{46.07}  & \textbf{63.72}  & \textbf{77.87}  &\textbf{ 23.75}  & \textbf{14.64}  & \textbf{25.60}  & \textbf{42.33}\\
				\bottomrule
			\end{tabular}
			\label{table2}
		}
	\end{table*}	
	
	\section{Experiments}
	In this section, we conduct many experiments to evaluate LogCL on four public datasets, and analyze the experimental results in detail.
	\subsection{Datasets and Baselines}
	\subsubsection{Datasets} To evaluate LogCL on entity prediction task, we adopt four benchmark datasets that are widely used for TKG extrapolation,  including  ICEWS14, ICEWS18, ICEWS05-15 \cite{boschee2015icews} and GDELT \cite{GDELT}. ICEWS14, ICEWS18 and ICEWS05-15 are subsets generated from Integrated Crisis Early Warning System (ICEWS)  \footnote{http://www.icews.com/}, which contains a large number of political events with specific timestamps. GDELT \footnote{https://www.gdeltproject.org/} is a subset generated from the Global Database of Events, Language datasets, which contains 20 of types events (e.g. Make a public statement, Appeal and consult). We follow the preprocessing strategy \cite{RE-NET,RE-GCN} to split all datasets into training, validation, and test sets with the proportions of 80\%/10\%/10\%. The statistics of all datasets are provided in Table \uppercase\expandafter{\romannumeral2}.
	\subsubsection{Baselines}
	To demonstrate the effectiveness of LogCL for TKG reasoning, we compare LogCL with 20 up-to-date KG reasoning methods that cover SKG and TKG  reasoning methods (including interpolation and extrapolation).  The SKG methods include DisMult \cite{DisMult}, ConvE \cite{Conv2D}, ComplEx \cite{CompIEx}, Conv-TransE \cite{Conv_TransE}, RotatE \cite{RotatE}. The interpolation methods include TTransE\cite{TTransE}, TA-DisMult \cite{TA-TransE}, De-SimIE\cite{DE-SimplE}, TNTComIEx\cite{TComplEX}, TANGO-Tucker \cite{TANGO}. The extrapolation methods include xERTE\cite{XERTE}, TITer\cite{TITer}, CyGNet\cite{CyGNet}, RE-NET\cite{RE-NET}, RE-GCN\cite{RE-GCN}, CEN\cite{CEN}, TiRGN\cite{TiRGN}, HisMatch \cite{Hismatch}, RETIA \cite{RETIA}, and CENET\cite{CENET}.
	
	\subsection{Experimental Settings}
	
	\subsubsection{Evaluation Metrics}
	We adopt two evaluation metrics: mean reciprocal rank (MRR) and Hits@k (k=1,3,10), which are widely used to evaluate the effectiveness of TKG reasoning methods. MRR is the average reciprocal value that is used to compute the ranks of the ground truth for all queries, and Hits@k represents the proportion of times that the true entity candidates appear in the top-k of the ranked candidates. 
	Some recent works\cite{TANGO,Hismatch} point out that the traditional static filtered setting is not suitable for extrapolation on TKG reasoning due to ignoring the time dimension of facts.  Actually,  only the facts occurring at the same time should be filtered. 
	Therefore, we report the experimental results with the time-aware filtered setting that is widely used in recent works, which only filters out the quadruples occurring at the query time. In addition, all experimental results on MRR and Hits@1/3/10  in this paper are reported as percentages.
	\subsubsection{Implementation Details}
	For all datasets, the embedding size $d$ is set to 200, the learning rate is set to 0.001 and the batch size is set as the number of quadruples in each timestamp. The parameters of LogCL are optimized by using adam \cite{Adam} during training.  The layer number of R-GCN on both local entity-aware attention recurrent encoder and global entity-aware attention encoder is set to 2 and the dropout rate for each layer is set to 0.2. The optimal local historical KG snapshots sequences lengths of ICEWS14, ICEWS18, ICEWS05-15, GDELT are set to 7, 7, 9 and 7,  respectively. We follow works\cite{RE-GCN,TiRGN,RETIA} that add static KG information on ICEWS14, ICEWS18 and ICEWS05-15 datasets. The prediction weight of all datasets is set  to 0.9.  The optimal  temperature coefficient of ICEWS14, ICEWS18, ICEWS05-15, GDELT are set to 0.03, 0.03, 0.07 and 0.07. For the decoder on all datasets, the number of kernels is set to 50, the kernel size is set to 2$\times$3, and the dropout rate is set to 0.2. 
	
	For SKG reasoning methods, the time dimension is removed on all TKG datasets. Some of the extrapolation baselines, including xERTE\cite{XERTE}, TANGO-Trucker\cite{TANGO}, TITer\cite{TITer}, REGCN\cite{RE-GCN}, CEN\cite{CEN}, TiRGN \cite{TiRGN} and RETIA\cite{RETIA}, publish their open source code, and their  results with the time-aware filter setting are reported  under the default parameters. The others are taken from the Hismathch \cite{Hismatch}. For fairness of comparison, results of CEN and RETIA  are reported under the offline setting that is adopted to other baselines. We also compare our LogCL with CEN and RETIA models under the online setting and results are reported in the following section \textit{H}.
	
	\begin{table*}[htbp]
		\caption{The ablation study results of MRR and Hits@1/3/10 on ICEWS14, ICEWS18 and ICEWS05-15 datasets.}
		\setlength\tabcolsep{3.5pt}
		\renewcommand{\arraystretch}{1.2}
		\centering
		\scalebox{1}{
			\begin{tabular}{lcccccccccccc}
				\toprule
				\multirow{2}{*}{Model}
				&\multicolumn{4}{c}{ICEWS14}&\multicolumn{4}{c}{ICEWS18}&\multicolumn{4}{c}{ICEWS05-15}\\
				\cmidrule(lr){2-5} \cmidrule(lr){6-9} \cmidrule(lr){10-13} 
				&MRR &Hits@1 &Hits@3 &Hits@10 &MRR &Hits@1 &Hits@3 &Hits@10 &MRR &Hits@1 &Hits@3 &Hits@10 \\
				\midrule
				LogCL &\textbf{48.87}  & \textbf{37.76}  & \textbf{54.71}  & \textbf{70.26}  & \textbf{35.67}  & \textbf{24.53}  & \textbf{40.32}  & \textbf{57.74}  & \textbf{57.04}  & \textbf{46.07}  & \textbf{63.72}  & \textbf{77.87}   \\
				LogCL-$\mathcal{G}$ &44.74  & 33.53  & 50.29  & 66.72  & 30.21  & 18.96  & 34.45  & 52.69  & 51.92  & 40.31  & 58.37  & 74.66  \\
				LogCL-$\mathcal{L}$  & 46.81  & 35.52  & 52.54  & 68.91  & 35.31  & 24.14  & 39.92  & 57.40  & 56.78  & 45.73  & 63.50  & 77.59  \\
				LogCL-w/o-eatt  &40.34  & 30.29  & 44.82  & 60.05  & 31.01  & 20.99  & 34.88  & 50.71  & 46.25  & 35.76  & 51.71  & 66.50  \\
				LogCL-$\mathcal{G}$-w/o-eatt  &38.61  & 29.04  & 42.79  & 57.42  & 27.83  & 18.56  & 31.11  & 46.30  & 41.40  & 31.20  & 46.44  & 60.93  \\
				LogCL-$\mathcal{L}$-w/o-eatt  &39.86  & 29.90  & 44.40  & 59.48  & 30.95  & 21.01  & 34.85  & 50.55  & 46.16  & 35.56  & 51.58  & 66.70  \\
				LogCL-w/o-cl  &46.84  & 35.62  & 52.44  & 69.05  & 35.32  & 24.04  & 40.00  & 57.62  & 56.85  & 45.83  & 63.61  & 77.81  \\
				\bottomrule
			\end{tabular}
			\label{table3}
		}
	\end{table*}
	
	\begin{figure*}[htbp]
		\centering
		\subfigure[The results of different intensity of noise in ICEWS14 dataset]{
			\begin{minipage}[t]{0.3\linewidth}
				\centering
				\includegraphics[width=1.05\linewidth]{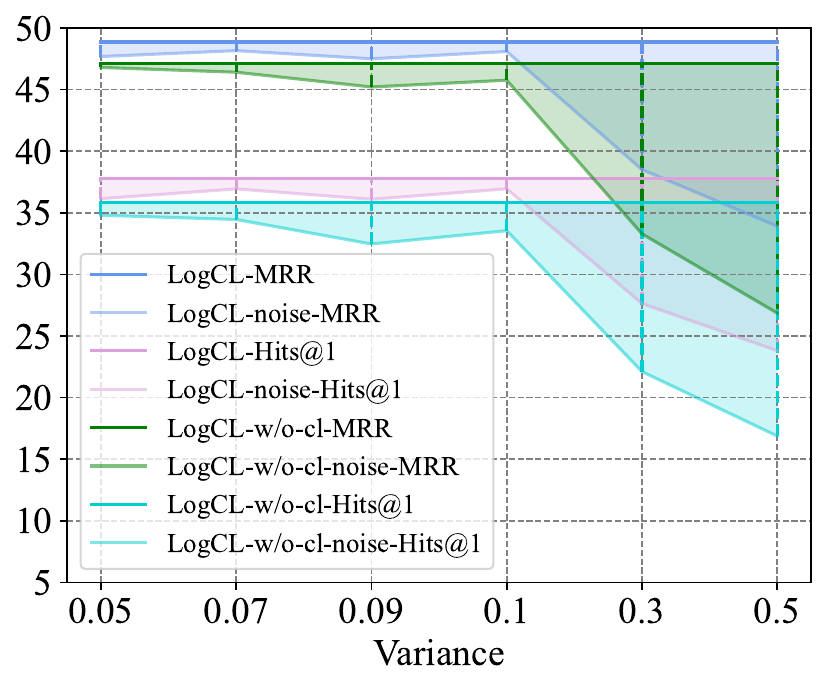}
			\end{minipage}
		}%
		\subfigure[The results of different intensity of noise in ICEWS18 dataset]{
			\begin{minipage}[t]{0.3\linewidth}
				\centering
				\includegraphics[width=1.05\linewidth]{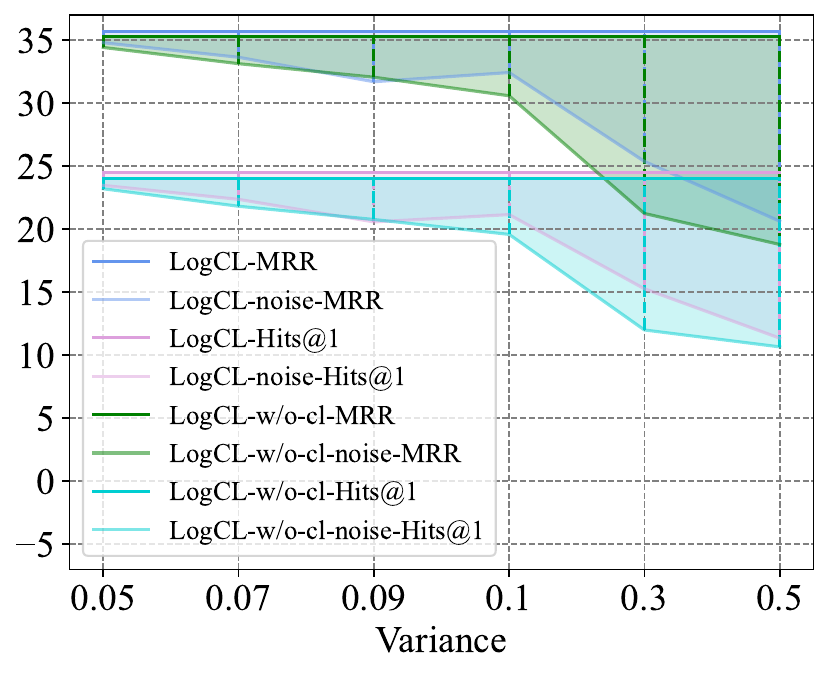}
			\end{minipage}
		}
		\subfigure[The results of different intensity of noise in ICEWS05-15 dataset]{
			\begin{minipage}[t]{0.3\linewidth}
				\centering
				\includegraphics[width=1.05\linewidth]{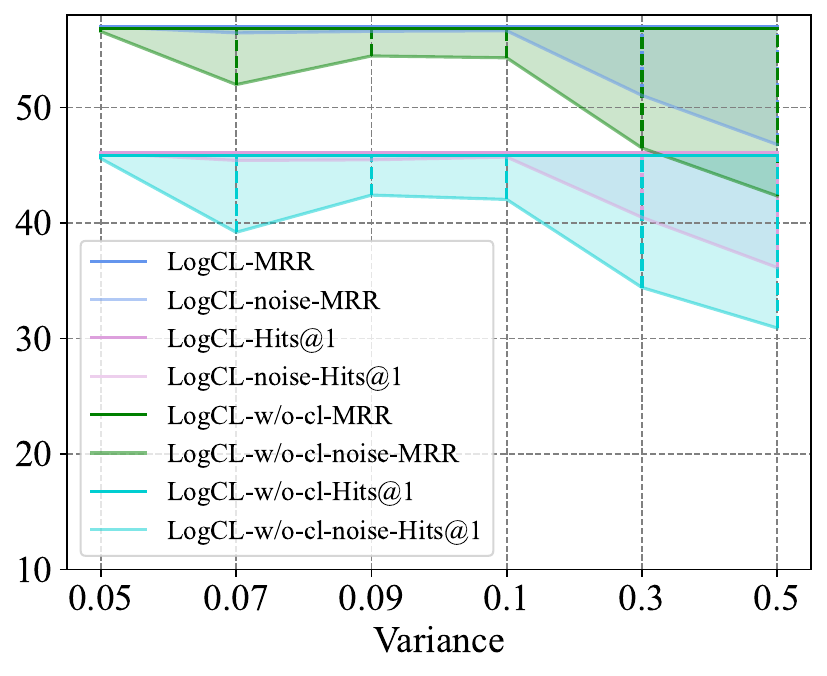}
			\end{minipage}
		}
		\caption{Study on the different intensity of noise on ICEWS14,ICEWS18 and ICEWS05-15 datasets on Hits@1 and MRR.}
		\label{Fig5}
	\end{figure*}
	
	\begin{figure}[htbp]
		\centering
		\subfigure[The results of different layers  on ICEWS14 dataset]{
			\begin{minipage}[t]{0.47\linewidth}
				\centering
				\includegraphics[width=1.05\linewidth]{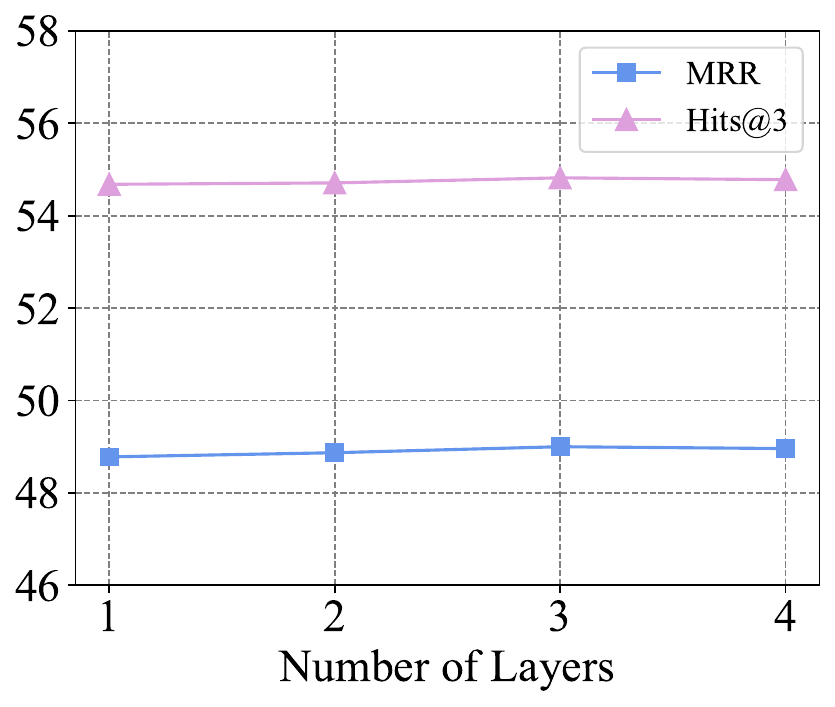}
			\end{minipage}
		}%
		\subfigure[The results of different layers on ICEWS18 dataset]{
			\begin{minipage}[t]{0.47\linewidth}
				\centering
				\includegraphics[width=1.05\linewidth]{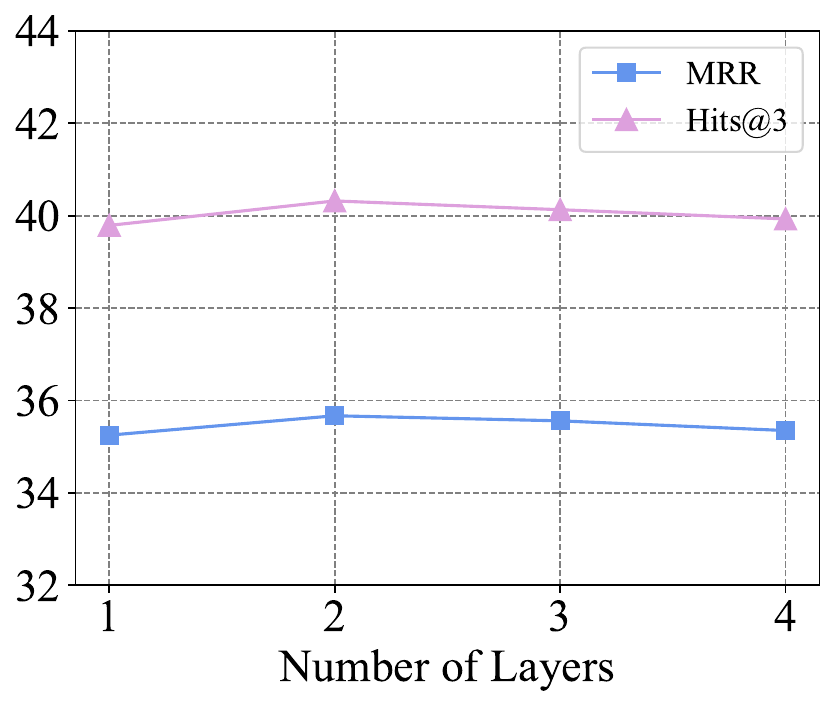}
			\end{minipage}
		}
		
		\caption{Study on the number of layers in R-GCN in the global entity-aware attention encoder  on ICEWS14 and ICEWS18 datasets.}
		
		\label{Fig8}
	\end{figure}
	
	\subsection{Results of TKG Extrapolation}
	The overall experimental results of LogCL and baselines on four benchmark datasets are reported in Table \uppercase\expandafter{\romannumeral3}. The best results are marked in bold, and the second-best results are reported using underlining. From the experimental results, we have the following observation:
	
	\begin{itemize}
		\item  The performance of LogCL consistently is better than the state-of-the-art methods on four benchmark datasets. Specifically, when comparing the second-best results, LogCL achieves the most significant improvements of 5.2\%, 4.9\%, 7.9\%, and 7.9\% in MRR on ICEWS14, ICEWS18, ICEWS05-15 and GDELT datasets. 
		\item For the state-of-the-art models Hismatch and TIRGN, although HisMatch also considers the local historical information related to the query, it fails to effectively consider the importance of the local historical fact information related to the query and does not consider the global historical information, resulting in weaker performance than LogCL. TiRGN takes into account both global and local historical information, but ignoring query-related historical information leads to lower performance than Hismath.
		Although CENET considers historical contrastive learning with our model LogCL, its performance is lower than LogCL due to the lack of evolutionary modeling of facts. In addition, CENET uses historical contrastive learning to enhance entities with less historical interactions by contrast, while we use contrastive learning from a local and global view to improve the robustness of the model.
		\item   An interesting phenomenon on all models is that the results on ICEWS14 and ICEWS18 datasets perform better  than  GDELT and ICEWS05-15 datasets. The reason for such phenomenon may be that more complex dynamic interactions among entities and relations exist in ICEWS18 and GDELT datasets. Such complex dynamic facts association are difficult to be modeled  and lead to the worse results on ICEWS18 and GDELT datasets. The excellent performance exhibited by LogCL on ICEWS18 and GDELT datasets demonstrates  the ability of LogCL to model complex temporal facts interactions.
		\item Extrapolation models on all datasets perform better than interpolation and static models, mainly because static models do not consider temporal information and are difficult to capture the dynamic changes of entities and relations. For interpolation models, they only focus on the completion of historical missing facts and lack the ability to model the evolution of entities and relationships over time and predict future unseen facts.
	\end{itemize}

	\begin{figure}[htbp]
	\centering
	\subfigure[The MRR results under different query contrast strategies]{
		\begin{minipage}[t]{0.47\linewidth}
			\centering
			\includegraphics[width=1.05\linewidth]{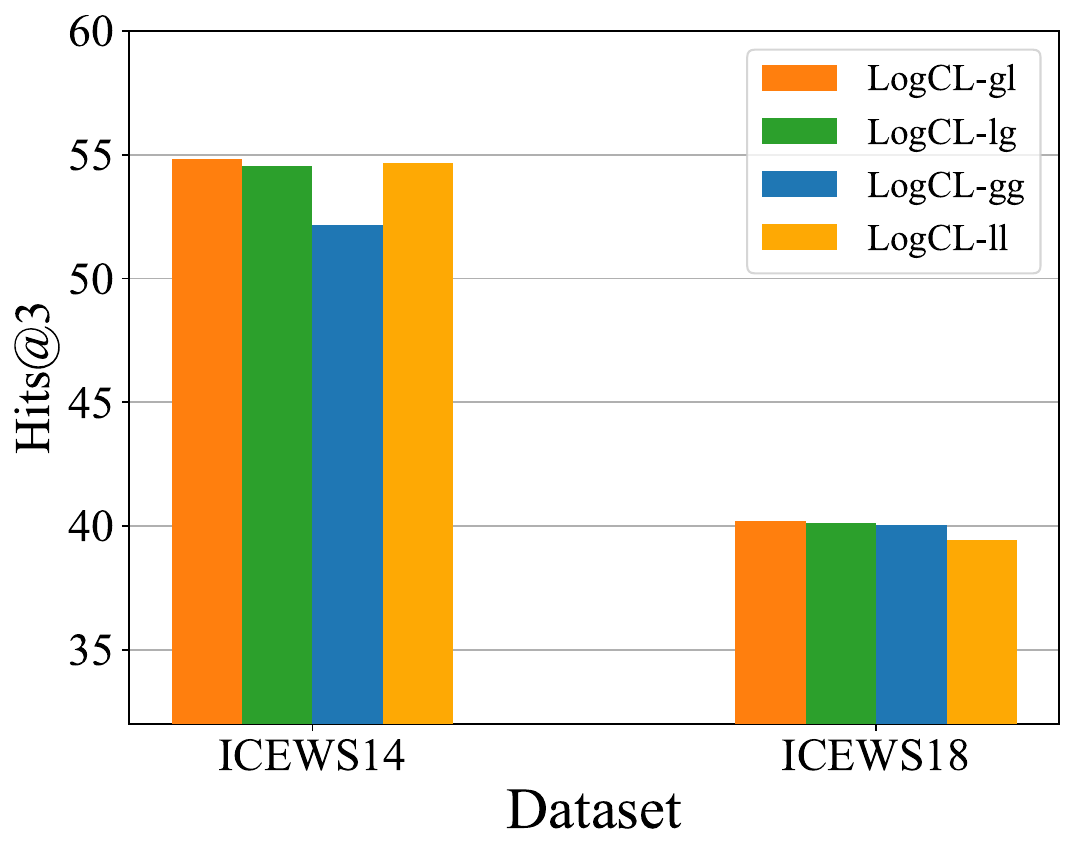}
		\end{minipage}
	}%
	\subfigure[The Hits@1 results under different query contrast strategies]{
		\begin{minipage}[t]{0.47\linewidth}
			\centering
			\includegraphics[width=1.05\linewidth]{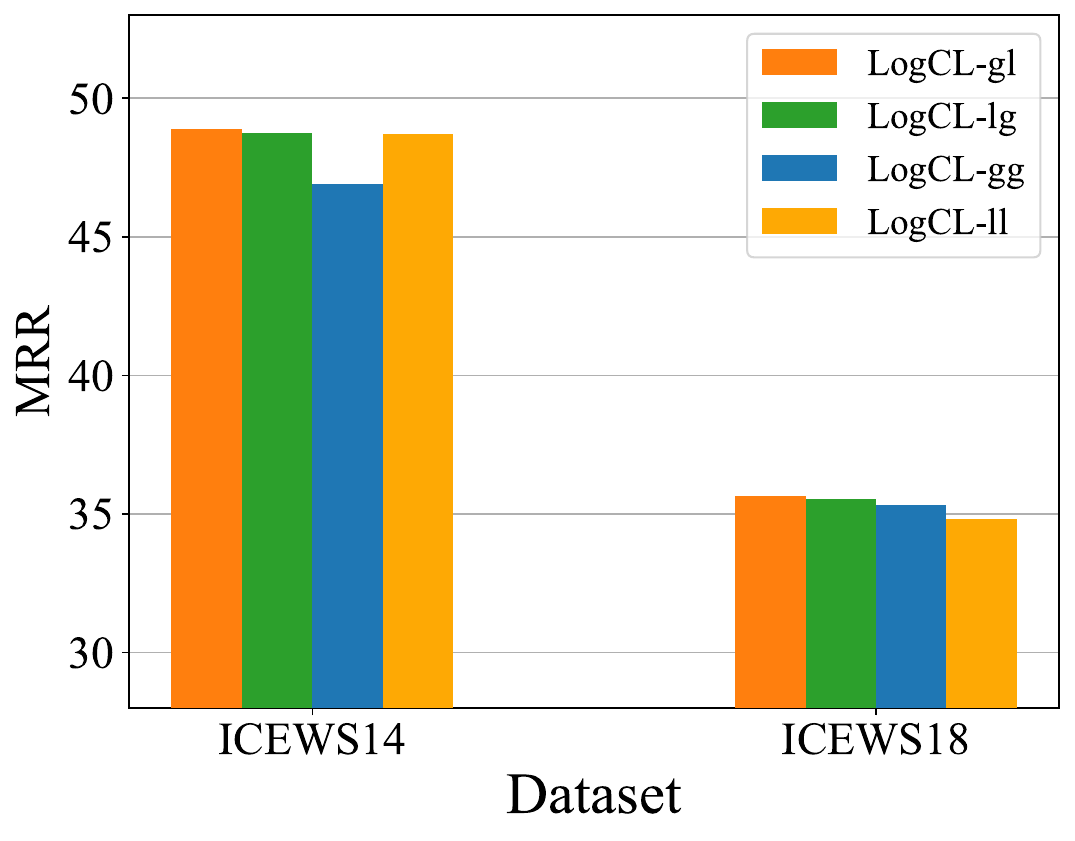}
		\end{minipage}
	}
	\caption{Study on different query contrast strategies on ICEWS14 and ICEWS18 datasets.}
	\label{Fig6}
\end{figure}

	\subsection{Ablation Study}
	To further better analyze each part of LogCL that contributes to the prediction results, we  conduct ablation studies on all datasets in terms of MRR and Hits@1/5/10. The results of the LogCL variants are presented in Table \uppercase\expandafter{\romannumeral4}.  
	\subsubsection{Impact of Global Entity-Aware Attention Encoder} The results denoted as LogCL-$\mathcal{L}$  in Table \uppercase\expandafter{\romannumeral4}  demonstrate the performance of LogCL without considering the global entity-aware attention encoder. It can be seen that the performances of   LogCL-$\mathcal{L}$ consistently performs poorly compared to the LogCL on ICEWS14, ICEWS18 and GDELT datasets, indicating that modeling the longer facts related to the queries is beneficial for TKG reasoning. We also conduct experiments to investigate the impact of varying the number of layers in the R-GCN in the global entity-aware attention encoder, and the results are shown in Fig. 6. It can be observed that demonstrate that the two-hop results are indeed slightly better than the one-hop results on both datasets. However, increasing the number of hops beyond two does not significantly improve performance on the ICEWS14 dataset and leads to decreased performance on the ICEWS18 dataset. A similar phenomenon also can be found in the local entity-aware attention recurrent encoder. 
	\subsubsection{Impact of Local Entity-Aware Attention Recurrent Encoder} LogCL-$\mathcal{G}$ in Table \uppercase\expandafter{\romannumeral4} is a variant of LogCL without modeling the recent local history at the adjacent timestamps.   
	More specifically, the local entity-aware attention recurrent encoder is removed in LogCL-$\mathcal{G}$.  The scores of
	all entities are obtained by using the representation of entities from the global entity-aware attention encoder and the static relation embedding as the representation of the query relation.  The results of LogCL-$\mathcal{G}$ in Table \uppercase\expandafter{\romannumeral4} can be observed that ignoring the local entity-aware attention recurrent encoder can generate a great impact on the performances. 
	Mainly because the recent local history contains rich information describing the behavior trends related to the query, which helps in the selection of the correct answer.
	Another observation is that the variant LogCL-$\mathcal{L}$  is superior to the variant LogCL-$\mathcal{G}$, indicating that modeling  the recent local entities and relations evolution is more effective than long historical information.

	\begin{figure}[htbp]
		\centering
		\subfigure[The results of different $\lambda$ on on ICEWS14 dataset]{
			\begin{minipage}[t]{0.47\linewidth}
				\centering
				\includegraphics[width=1.05\linewidth]{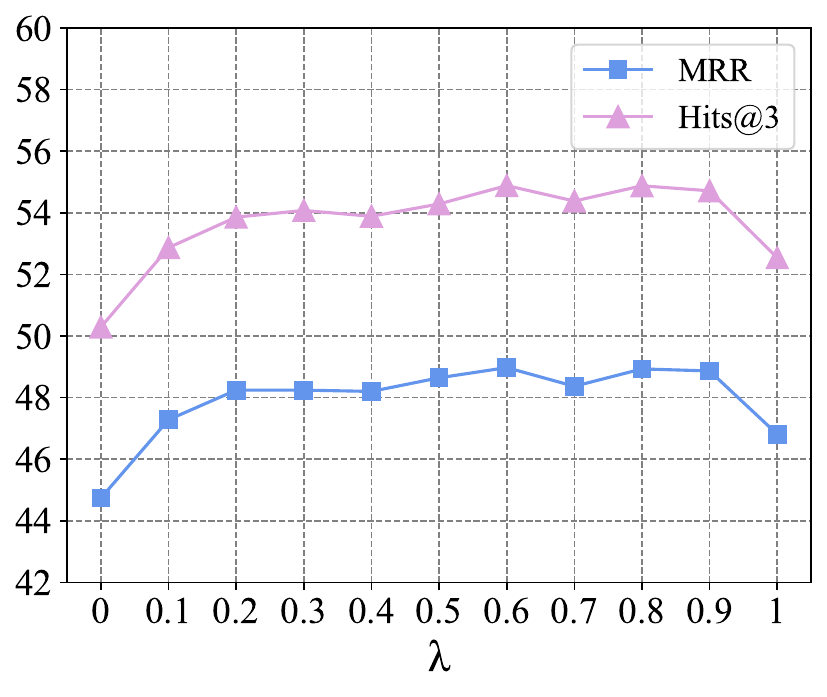}
			\end{minipage}
		}%
		\subfigure[The results of different $\lambda$ on ICEWS18 dataset]{
			\begin{minipage}[t]{0.47\linewidth}
				\centering
				\includegraphics[width=1.05\linewidth]{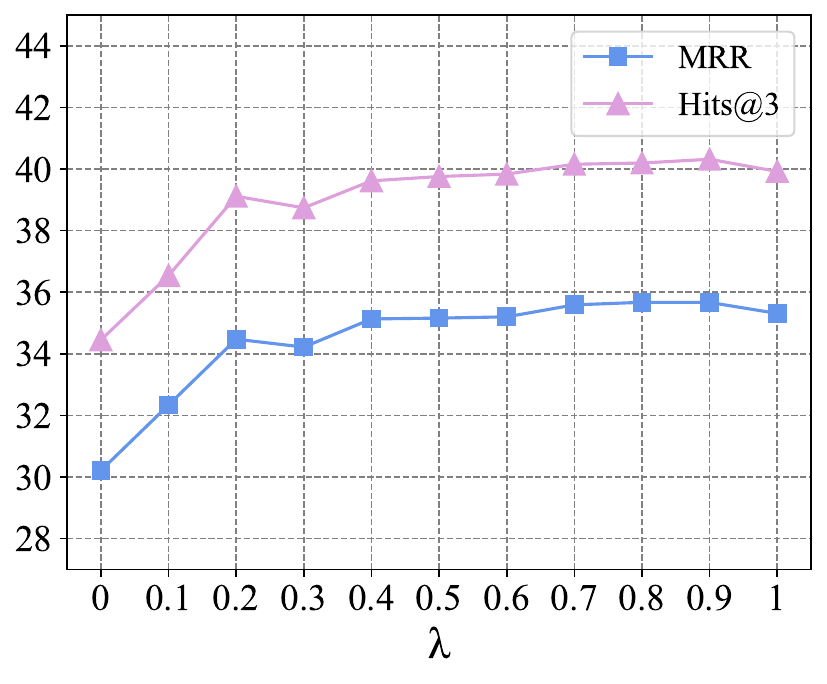}
			\end{minipage}
		}
		\caption{Study on the parameter $\lambda$ on ICEWS14 and ICEWS18 datasets.}
		\label{Fig6}
	\end{figure}

	\begin{figure}[htbp]
		\centering
		\subfigure[The results of different temperature coefficient $\tau$ on ICEWS14 dataset]{
			\begin{minipage}[t]{0.47\linewidth}
				\centering
				\includegraphics[width=1.05\linewidth]{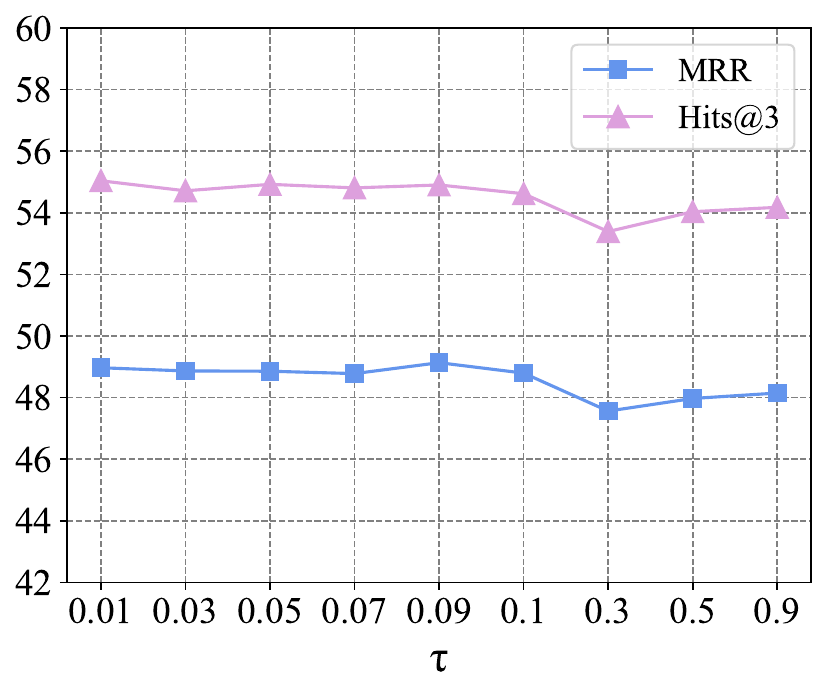}
			\end{minipage}
		}%
		\subfigure[The results of different temperature coefficient $\tau$ on ICEWS18 dataset]{
			\begin{minipage}[t]{0.47\linewidth}
				\centering
				\includegraphics[width=1.05\linewidth]{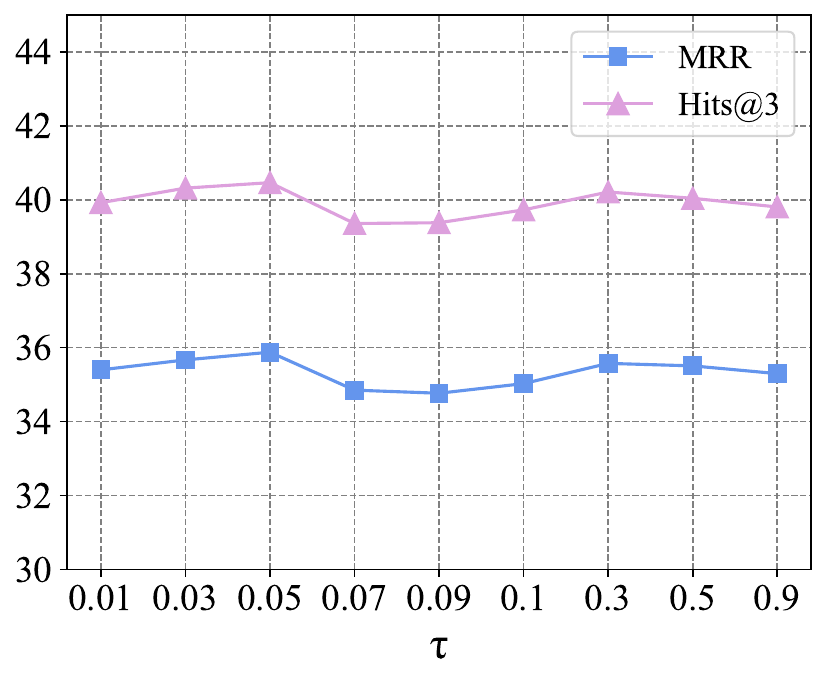}
			\end{minipage}
		}
		
		\caption{Study on the temperature coefficient $\tau$ in ICEWS14 and ICEWS18 datasets.}

		\label{Fig8}
	\end{figure}
	
	\subsubsection{Impact of Entity-Aware Attention} 
	To demonstrate how the entity-aware attention contributes to the prediction results, the entity-aware attention is removed in the local entity-aware attention recurrent encoder and global entity-aware attention encoder, which is denoted as "-w/o-eatt" in Table \uppercase\expandafter{\romannumeral4}. It can be intuitively observed  that  the performance degradation is great after removing the entity-aware attention on ICEWS14, ICEWS18 and ICEWS05-15 datasets. The main reason is that the entity attention mechanism can effectively capture the facts that the local KG snapshot and global historical subgraph are related to the query.
	
	\subsubsection{Impact of Contrastive Learning} 
	LogCL-w/o-cl denotes a variant of LogCL that removes the local-global contrastive learning. It can be observed that the performance of LogCL-w/o-cl is worse than that of LogCL, especially on the ICEWS14 dataset. It is because contrastive learning plays a role in guiding LogCL to fuse different historical characteristics.  We also conduct experiments to investigate the impact of four query contrast strategies. The results are presented in Fig. 7. LogCL-gl, LogCL-lg, LogCL-gg, and LogCL-ll denote variants using four different query losses, respectively. It can be observed that LogCL-gl and Log-lg demonstrate slightly superior performance compared to LogCL-gg and LogCL-ll. This suggests that contrastive training by emphasizing the distinction between local and global query representations yields greater benefits.
	
	To further analyze the ability of the local-global query contrast module of LogCL to resist noise interference, we conduct experiments by adding Gaussian noise with different intensities to LogCL and LogCL-w/o-cl. Here, Gaussian noise generated by the Gaussian distribution is added to the entity representation as the initial input of the model. As our primary objective is to predict future Gaussian entities, we solely introduce noise to the entities and not the relations. We control the intensity of Gaussian noises  by varying the variance in the Gaussian formula. The experimental results are shown in Fig. 5. When faced with the same intensity of noise, a larger span of Y-axis values on the same color indicates a larger performance degradation. We can intuitively observe that the LogCL model always performs better than the LogCL-w/o-cl model  under the same intensity of noise on ICEWS14, ICEWS18 and ICEWS05-15 datasets.  This shows that local-global contrastive learning can effectively improve the anti-noise ability of the model. In addition, we also observe another interesting phenomenon, although the performance of LogCL model and LogCL-w/o-cl model gradually decreases with the increase of noise intensity, it is worth noting that the performance of LogCL model decreases less than that of LogCL-w/o-cl model when facing stronger noise, especially on ICEWS05-15 dataset. LogCL shows relatively strong performance even when subjected to strong noise interference.
	
	\begin{table}[htbp]
		\caption{The ablation experiment results of MRR and Hits@1 on ICEWS14, ICEWS18 and ICEWS05-15 datasets.}
		\setlength\tabcolsep{3pt}
		\renewcommand{\arraystretch}{1.2}
		\centering
		\scalebox{0.95}{
			\begin{tabular}{lllllll}
				\toprule
				\multirow{2}{*}{Model}
				&\multicolumn{2}{c}{ICEWS14}&\multicolumn{2}{c}{ICEWS18}&\multicolumn{2}{c}{ICEWS05-15}\\
				\cmidrule(lr){2-3} \cmidrule(lr){4-5} \cmidrule(lr){6-7} 
				&MRR &Hits@1 &MRR &Hits@1 &MRR &Hits@1\\
				\midrule
				LogCL (RGCN)  &48.87 &\textbf{37.76} &35.67 &\textbf{24.53} &\textbf{57.04} &\textbf{46.07}  \\
				LogCL (CompGCN-sub) &\textbf{49.25} &36.84 &35.33 &24.26 &56.93 &45.92   \\
				LogCL (CompGCN-mult) & 47.92	& 36.85	&35.32	&24.05	&56.40	& 45.46	 \\
				LogCL (KBAT)  &48.46  &37.17  &\textbf{35.70}  &24.41  &56.01  &45.14  \\
				\bottomrule
			\end{tabular}
			\label{table3}
		}
	\end{table}
	
	\begin{figure}[htbp]
		\centering
		\subfigure[The MRR results under online learning]{
			\begin{minipage}[t]{0.47\linewidth}
				\centering
				\includegraphics[width=1.05\linewidth]{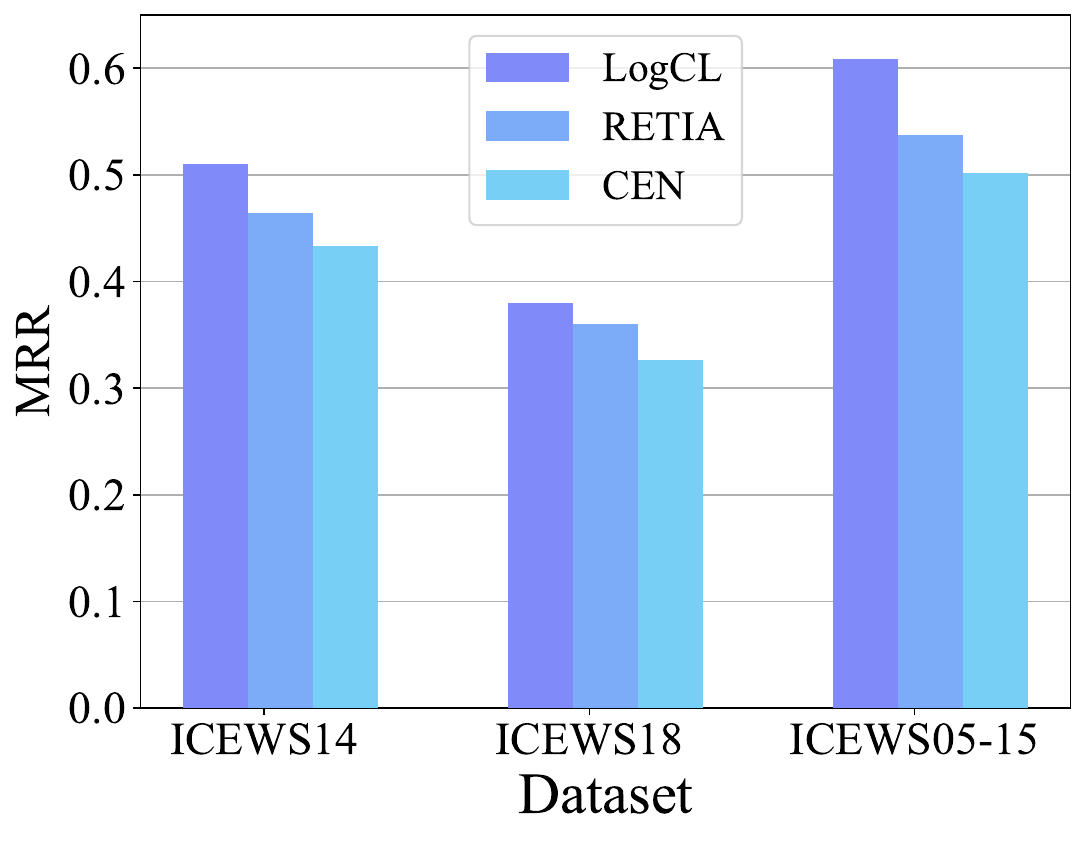}
			\end{minipage}
		}%
		\subfigure[The Hits@1 results under online learning]{
			\begin{minipage}[t]{0.47\linewidth}
				\centering
				\includegraphics[width=1.05\linewidth]{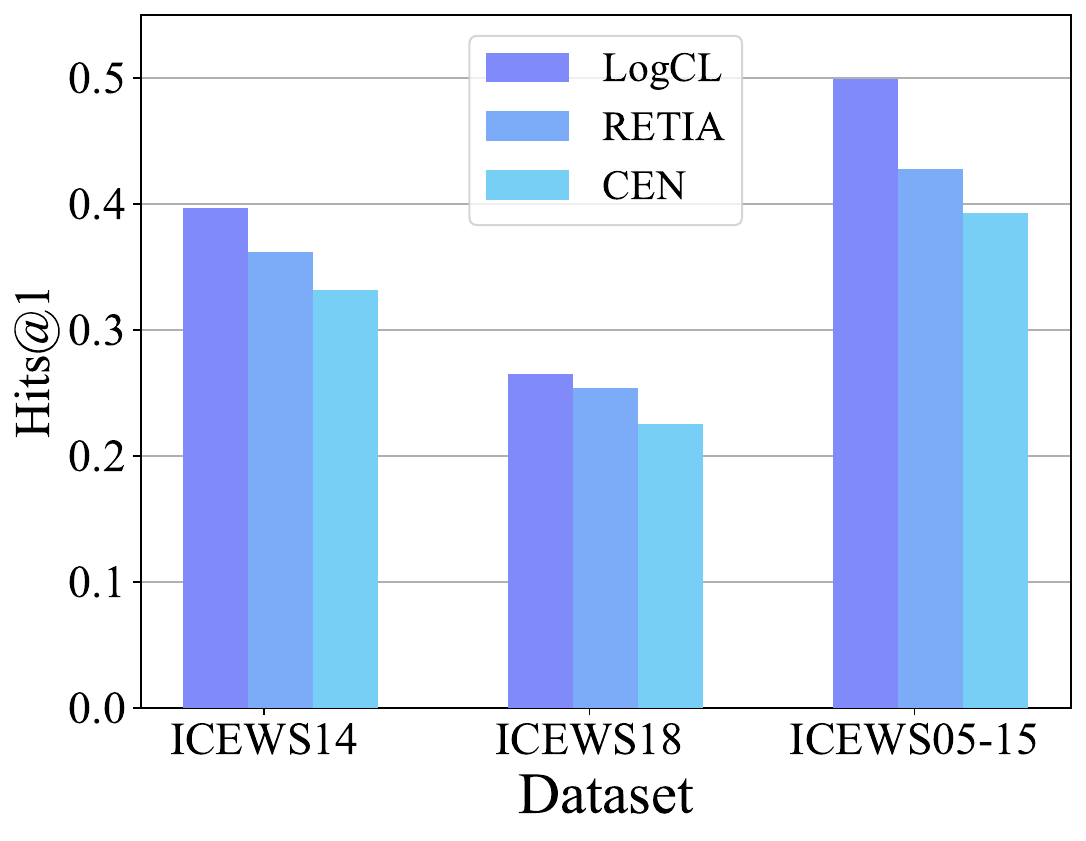}
			\end{minipage}
		}
		\caption{Study on online training on ICEWS14, ICEWS18 and ICEWS05-15 datasets.}
		\label{Fig6}
	\end{figure}
	
	\begin{table*}[htbp]
		\centering
		\caption{Two examples of top 5 predictions for the given queries on ICEWS14 dataset}
		\setlength\tabcolsep{2.5pt}
		\renewcommand{\arraystretch}{1.2}
		\scalebox{1}{
			\begin{tabular}{llll}
				\toprule
				{Query and Answer} & LogCL  & LogCL-w/o-eatt & LogCL-w/o-cl  \\
				\midrule
				\multirow{5}[2]{*}{ \makecell[l]{\textbf{Query:} \\( China, Sign\_\\formal\_agreement,\\ ?, 2014-12-4 )\\\textbf{Answer:}\\South\_Africa }} & \textbf{South\_Africa} \quad \textbf{0.6136} &  Kazakhstan \quad 0.109 & \textbf{South\_Africa} \quad \textbf{0.501} \\
				& Ashraf\_Ghani\_Ahmadzai\quad	0.088 & France \quad	0.083 & Ashraf Ghani Ahmadzai \quad 0.244\\ 
				& Malaysia \quad	0.084 & South\_Korea \quad 0.076 & South Korea \quad	0.058\\ 
				& South\_Korea \quad 0.045 & Vietnama \quad 0.069 & Malaysia \quad	0.050\\
				& European\_Parliament \quad	0.043 & Iraq \quad	0.049 & European\_Parliament \quad 0.029\\
				\midrule
				\multirow{5}[2]{*}{ \makecell[l]{\textbf{Query:} \\( Iran, Engage\_in\_\\diplomatic\_cooperation,\\ ?, 2014-11-30 )\\\textbf{Answer:}\\Oman } } & \textbf{Oman} \quad \textbf{0.687} & \textbf{Oman} \quad \textbf{0.1098} & \textbf{Oman} \quad \textbf{0.668}\\
				& Portugal \quad 0.122 & Iraq \quad 0.1095 &Portugal \quad 0.129\\
				& Guinea \quad 0.077 & Food\_and\_Agriculture\_Organization \quad 0.0813 &Iraq \quad 0.045\\
				& Qatqr \quad 0.034 & China \quad 0.0562 &Qatar\quad		0.041\\
				& China \quad 0.021 & Tajikistan \quad 0.0434 &Guinea \quad	0.012\\
				\bottomrule
		\end{tabular}}%
		\label{table5}%
	\end{table*}%
	
	\begin{table}[htbp]
		\caption{The results of the two-phase propagation on ICEWS14 , ICEWS18 and ICEWS05-15 datasets.}
		\setlength\tabcolsep{3pt}
		\renewcommand{\arraystretch}{1.2}
		\centering
		\scalebox{0.98}{
			\begin{tabular}{lllllll}
				\toprule
				\multirow{2}{*}{Model}
				&\multicolumn{2}{c}{ICEWS14}&\multicolumn{2}{c}{ICEWS18}&\multicolumn{2}{c}{ICEWS05-15}\\
				\cmidrule(lr){2-3} \cmidrule(lr){4-5} \cmidrule(lr){6-7} 
				&MRR &Hits@1 &MRR &Hits@1 &MRR &Hits@1\\
				\midrule
				LogCL  &48.87 &37.76 &35.67 &24.53 & 57.04 & 46.07 \\
				LogCL-$\mathcal{FP}$ &\textbf{50.69} &\textbf{39.66} &\textbf{37.38} &\textbf{25.96} &\textbf{58.69} &\textbf{47.79}   \\
				LogCL-$\mathcal{SP}$  & 47.04	& 35.87	&33.89	&22.97	&55.38	& 44.34	 \\
				\bottomrule
			\end{tabular}
			\label{table3}
		}
	\end{table}
	\subsection{Sensitivity Analysis}
	In this section, we conduct experiments on ICEWS14 and 18 datasets to further analyze the impact of parameters in LogCL, including the parameter $\lambda$ in the prediction part, the temperature coefficient $\tau$ in the local-global query contrast module, and GNN aggregation encoder.
	
	\subsubsection{Analysis of  the Parameter $\lambda$ }
	To explore the parameter $\lambda$ that is used to trade off global and local representations of entities, we conduct experiments with different values of $\lambda$ from 0 to 1 with other optimal hyperparameters fixed. The results of MRR and Hist@3 are shown in Fig.8. A larger value of $\lambda$ indicates a higher proportion of the local entity-aware attention recurrent encoder.
	It can be seen that the performance of LogCL shows a trend of first upward and then falling as $\lambda$ increases on ICEWS14 and ICEWS18 datasets, which indicates that only considering local historical information or global historical information cannot effectively predict the results. This trend further illustrates that an appropriate parameter is important for combining local and global historical patterns.
	
	\subsubsection{Analysis of Temperature Coefficient $\tau$}
	We perform a batch of experiments with different temperature coefficients with other optimal hyperparameters fixed. The results of MRR and Hits@3 are shown in Fig.9. Different datasets are affected by the temperature coefficient differently, and choosing the appropriate temperature coefficient is helpful for TKG reasoning.
	
	\subsection{Study On Different GNN Aggregation }
	To further study the impact of different kinds of GNNs in the local entity-aware attention recurrent encoder and the global entity-aware attention encoder, R-GCN is replaced in these two encoders with CompGCN\cite{CompGCN} and KBGAT\cite{KBGAT}. The MRR and Hits@1 results of ICEWS14, ICEWS18, and ICEWS05-15 are reported in Table \uppercase\expandafter{\romannumeral5}. It can be seen that LogCL (RGCN) gets the best results on ICEWS05-15 dataset and shows strong performance in ICEWS14 and ICEWS18 datasets.

		\subsection{Study On the Two-Phase Propagation }
		To investigate the role of  the two-phase propagation in our LogCL, we manage experiments to evaluate the results of two forward propagations separately. Log-$\mathcal{FP}$ denotes a variant that predicts the object entities on the origin query set and only performs the first phase propagation. Log-$\mathcal{SP}$ that predicts the subjected entities on the reverse query set and only considers the second phase propagation. The results are shown in Table \uppercase\expandafter{\romannumeral7}. It can be seen that the performance of Log-$\mathcal{FP}$ is superior to the  Log-$\mathcal{SP}$. 
		The reason behind this phenomenon is that the dataset composed of inverse relations may introduce a certain bias, which subsequently impacts the performance of the model. In contrast, the original dataset provides a more comprehensive reflection of the real relations between entities, resulting in better performance in the experimental results.
	
	\subsection{Study On the Online Training}
	Since the evolutionary pattern changes with emerging facts, exploring how to adjust the model to emerging facts is crucial. Following the CEN \cite{CEN} and RETIA \cite{RETIA}, we perform a batch of experiments to learn emerging facts on ICEWS14, ICEWS18 and ICEWS05-15 datasets under the online setting.  The results are shown in the Fig.10. The results of CEN, RETIA and LogCL under the online setting outperform the results in Table \uppercase\expandafter{\romannumeral3} under the offline setting. It is because that historical facts in the test set are updated under the online setting.  In addition, LogCL achieves greater improvements than the CEN and RETIA under the online setting. This proves that LogCL can effectively solve domain obstacles such as time-varying problems.
	
	\subsection{Case Study}

		We provide the case study for LogCL, LogCL-w/o-eatt, and LogCL-w/o-cl on ICEWS14 dataset.  Two queries with the top-5 prediction entities and scores are reported in Table \uppercase\expandafter{\romannumeral6}. 
		
		For the query (China, Sign\_formal\_agreement, ?, 2014-12-4), we can observe that LogCL and logCL-w/o-cl can predict the correct answer, while the top-5 results given by LogCL-w/o-eatt do not contain the correct answer. For the query (Iran, Engage\_in\_diplomatic\_cooperation, ?, 2014-11-30), although LogCL,  LogCL-w/o-eatt, and LogCL-w/o-eatt can give answers prepared and the answers are top-1, the prediction scores of LogCl and LogCL-w/o-eatt for correct answers are much higher than that of LogCL-w/o-eatt. The reason for the above phenomenon is that the reasoning of LogCL-w/o-eatt primarily relies on historical facts that are in close proximity in time and weakly associated with the query. As a result, LogCL-w/o-eatt fails to capture the historical facts that have a strong relevance to the query, leading to lower accurate prediction results.
		These two cases can further prove the effectiveness of the entity-aware attention mechanism and the strong reasoning ability of LogCL.

	\section{Conclusion}
	In this paper, we propose a novel TKG reasoning model, namely LogCL, which uses contrastive learning to better guide the model to integrate local and global historical information, thereby improving the robustness of the model.   
	We propose a local entity-aware attention recurrent encoder, which effectively captures the importance of query-related historical information in each KG snapshot at the most recent time.
	We propose a global entity-aware attention encoder, which learns the importance of the global historical information in the history subgraph that is built by sampling the global historical facts based on queries.
	We design a local-global query contrast module to focus on the important common features of the local encoder and global encoder, so as to enhance the anti-noise ability of LogCL.
	Extensive experiments on four public datasets demonstrate that our proposed LogCL performs significant improvements and has a more robust performance than the state-of-the-art baselines.
	
	\bibliographystyle{IEEEtran}
	\bibliography{ICDE2024_LogCL}

\end{document}